\documentclass[10pt,twocolumn,letterpaper]{article}

\usepackage[pagenumbers]{cvpr}              %

\usepackage{amsmath}
\usepackage{bbm}

\newcommand{\R}{I\!\!R} %

\newcommand{\visionfeature}{\mathbf{x}}
\newcommand{\languagefeature}{\mathbf{y}}

\newcommand{\visionkernel}{k_x}
\newcommand{\languagekernel}{k_y}
\newcommand{\visionpairwise}{\mathbf{X}}
\newcommand{\languagepairwise}{\mathbf{Y}}

\newcommand{\bigO}{\mathcal{O}}

\DeclareMathOperator*{\argmin}{arg\,min}
\DeclareMathOperator*{\argmax}{arg\,max}

\DeclareMathOperator{\topk}{top}
\DeclareMathOperator{\tr}{tr}

\newcommand*{\inparagraph}[1]{\noindent\textbf{#1}\hspace{0.5em}}

\newcommand{\hg}{Hahn-Grant\xspace}
\newcommand{\gw}{Gromov-Wasserstein\xspace}
\newcommand{\gwshort}{$\mathcal{GW}$\xspace}

\usepackage{tikz}
\usetikzlibrary{positioning, calc, patterns, shapes.arrows, shapes.geometric, fit, arrows, backgrounds, shapes.callouts}
\usepackage{pgfplots}
\usepgfplotslibrary{groupplots,dateplot}
\pgfplotsset{compat=newest}

\usetikzlibrary{external}
\definecolor{TUMBlue}{HTML}{0065BD}
\definecolor{TUMSecondaryBlue}{HTML}{005293}
\definecolor{TUMSecondaryBlue2}{HTML}{003359}
\definecolor{TUMBlack}{HTML}{000000}
\definecolor{TUMWhite}{HTML}{FFFFFF}
\definecolor{TUMDarkGray}{HTML}{333333}
\definecolor{TUMGray}{HTML}{808080}
\definecolor{TUMLightGray}{HTML}{CCCCC6}
\definecolor{TUMAccentGray}{HTML}{DAD7CB}
\definecolor{TUMAccentOrange}{HTML}{E37222}
\definecolor{TUMAccentGreen}{HTML}{A2AD00}
\definecolor{TUMAccentLightBlue}{HTML}{98C6EA}
\definecolor{TUMAccentBlue}{HTML}{64A0C8}

\definecolor{TUMYellow}{RGB}{254, 215, 2}
\definecolor{TUMPink}{RGB}{181, 92, 165}
\definecolor{TUMPurple}{RGB}{143, 129, 234}
\definecolor{TUMRed}{RGB}{234, 114, 55}

\usepackage{algorithm}
\usepackage[noend]{algorithmic}

\usepackage{multirow}
\usepackage{pdflscape}
\usepackage{array}
\usepackage{setspace}
\usepackage{tabularx}
\usepackage{siunitx}
\newcolumntype{L}[1]{>{\raggedright\let\newline\\\arraybackslash\hspace{0pt}}m{#1}}
\newcolumntype{C}[1]{>{\centering\let\newline\\\arraybackslash\hspace{0pt}}m{#1}}
\newcolumntype{R}[1]{>{\raggedleft\let\newline\\\arraybackslash\hspace{0pt}}m{#1}}

\usepackage{mathtools}

\DeclarePairedDelimiter\floor{\lfloor}{\rfloor}

\newcommand{\qlinc}[1]{\scriptsize\textcolor{TUMAccentGreen}{$\uparrow \phantom{-}#1$}}
\newcommand{\qldec}[1]{\scriptsize\textcolor{TUMRed}{$\downarrow #1$}}
\newcommand{\qsinc}[1]{\scriptsize\textcolor{TUMRed}{$\uparrow \phantom{-}#1$}}
\newcommand{\qsdec}[1]{\scriptsize\textcolor{TUMAccentGreen}{$\downarrow #1$}}
\newcommand{\qeq}[1]{\scriptsize{$=#1$}}

\usepackage[accsupp]{axessibility}
\usepackage{nccmath}
\AtBeginEnvironment{align}{\useshortskip}

\definecolor{cvprblue}{rgb}{0.21,0.49,0.74}
\usepackage[pagebackref,breaklinks,colorlinks,allcolors=cvprblue]{hyperref}

\def\paperID{125} %
\def\confName{CVPR}
\def\confYear{2025}

\usepackage{fancyhdr}
\usepackage{setspace}

\title{It's a (Blind) Match! \\Towards Vision-Language Correspondence without Parallel Data}

\author{Dominik Schnaus \hspace{1cm} Nikita Araslanov$^\dagger$ \hspace{1cm} Daniel Cremers$^\dagger$ \\[2mm]
TU Munich \hspace{1cm} Munich Center for Machine Learning\\[1mm]
{\small \textbf{Project page:} {\tt\href{https://dominik-schnaus.github.io/itsamatch/}{dominik-schnaus.github.io/itsamatch}} \hspace{1cm} $^\dagger$ equal advising}%
}

\begin{document}
\maketitle
\thispagestyle{fancy}

\begin{abstract}
   The platonic representation hypothesis suggests that vision and language embeddings become more homogeneous as model and dataset sizes increase. In particular, pairwise distances within each modality become more similar. 
    This suggests that as foundation models mature, it may become possible to match vision and language embeddings in a fully unsupervised fashion, \ie without parallel data.
    We present the first feasibility study, and investigate conformity of existing vision and language foundation models in the context of unsupervised, or ``blind'', matching. 
    First, we formulate unsupervised matching as a quadratic assignment problem and introduce a novel heuristic that outperforms previous solvers. We also develop a technique to find optimal matching problems, for which a non-trivial match is very likely. Second, we conduct an extensive study deploying a range of vision and language models on four datasets. Our analysis reveals that for many problem instances, vision and language representations can be indeed matched without supervision.
    This finding opens up the exciting possibility of embedding semantic knowledge into other modalities virtually annotation-free.
    As a proof of concept, we showcase an unsupervised classifier, which achieves non-trivial classification accuracy without any image-text annotation.
    
\end{abstract}
    
\section{Introduction}

\begin{figure}
    \centering
    \includegraphics{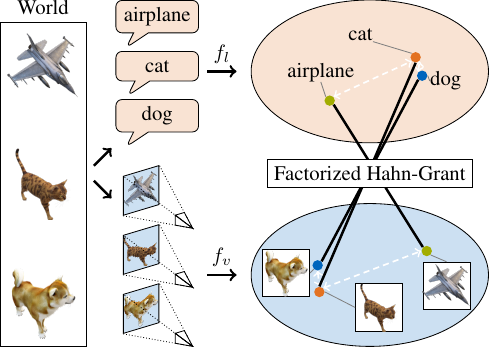}
    \caption{\textbf{Blind matching of vision and language:} Text and images are both abstractions of the same underlying world. Vision and language encoders $f_v$ and $f_l$ learn similar pairwise relations between concepts, \eg ``cat'' is closer to ``dog'' than to ``airplane''. We exploit these pairwise relations in our factorized Hahn-Grant solver to find valid correspondences between vision and language without any parallel data.}
    \label{fig:teaser}
    \vspace{-0.5em}
\end{figure}

Vision and language foundation models are changing the research landscape in computer vision and natural language processing. %
On the one hand, computer vision has widely adopted deep image representations, such as DINO~\cite{Caron:2021:EPS}, across virtually all applications, including semantic understanding \cite{HamiltonZHSF22} and 3D reconstruction \cite{Wu:2023:MPL}.
On the other hand, large language models (LLMs) \cite[\eg,][]{Devlin:2019:BER} are already excelling at imitating natural language, performing comparably or better than humans in language translation and basic reasoning \cite[\eg, see][for a survey]{Zhao:2023:ASL}.
Yet, we gravely lack in understanding the conformity between these models. %
Previous work offers some analysis under the lens of a single modality or in the cross-modal scenario with focus on label efficiency \cite{Norelli:2023:ASI,Moschella:2023:RRE}.
In particular, the latter body of work pursued dense alignment between vision and language embeddings from a few correspondences.
Taking this scenario to the extreme, we study ``blind'' matching, illustrated in \cref{fig:teaser}.
Specifically, we aim to find vision-language correspondence without \emph{any} parallel data.

Vision and language representations may seem incompatible at first. Indeed, there are significant differences in the training process, data distribution, and model architectures. 
However, the platonic representation hypothesis~\cite{huh2024platonic} posits that textual and visual representations of the same concepts converge to geometrically similar latent spaces. 
It asserts that models trained on large, diverse datasets encounter the same phenomena and, more importantly, the same relationships between the phenomena.
As \cref{fig:teaser} conceptually illustrates, existing vision and language models already exhibit a high degree of vision-language compatibility.
As \citet{huh2024platonic} also show, the alignment grows with larger sizes of models and training corpora.

Capitalizing on the platonic representation hypothesis, we study vision-language alignment from a completely unsupervised perspective.
Developing techniques towards such an analysis is valuable for two reasons.
\emph{First}, it will provide us with tools to study the conformity between vision and language models also on the abundance of unlabelled data.
\emph{Second}, establishing correspondence between semantic concepts (encapsulated by text) and images paves the way for purely unsupervised visual recognition.
In fact, our study culminates in an intriguing proof-of-concept application, \emph{unsupervised classifiers}, which perform (semantic) image recognition without any paired data.

\inparagraph{Contributions.}
As our \emph{technical} contribution to the study of blind vision-language alignment, we first formulate the task as a quadratic assignment problem (QAP).
This formulation requires only pairwise distances within each modality as the input.
We then discuss and analyze the metrics behind the pairwise relationships. %
Moving forward, we revisit and extend a solver for QAPs, exploiting a specialized heuristic towards memory-efficient and accurate vision-language matching.
We thoroughly compare our solver extension with previous heuristics for QAPs and vision-language matching, establishing that our solver leads to better optima and provides tighter bounds. %
Lastly, we define optimal criteria for finding suitable subsets of matching problems as a p-dispersion-sum problem~\cite{kuby1987programming}.

As our main \emph{empirical} contribution, we conduct a large-scale study involving a total of 33 vision and 27 language models on four datasets.
Our study reveals that despite the complexity of the underlying optimization problem, we can establish non-trivial solutions in some of the vision-language assignment problems across varying problem sizes.
Intriguingly, this finding enables \emph{unsupervised classification}: classifying images into semantic categories without any parallel supervision.

\section{Related work}
\label{sec:related}

\inparagraph{Vision and language foundation models.}
Unsupervised learning has enabled large-scale pre-training, both in the vision \cite{Chen:2020:ASF,Caron:2021:EPS,He:2022:MAA,Oquab:2024:DIN} and language domains \cite{Devlin:2019:BER,Reimers:2019:SBE,Chowdhery:2023:PLM}.
While semantics are inherent to language, the emergence of semantics in unsupervised vision models is yet not well-understood \cite{Caron:2021:EPS}.
Unlike unimodal foundation models, joint vision-text encoders, such as CLIP \cite{Radford:2021:LTV} and ALBEF \cite{Li:2021:ABF}, as well as generative text-to-image models \cite{Ramesh:2021:ZST,Saharia:2022:PTI}, leverage a shared vision-language embedding space \cite{Ramesh:2022:HTC}. %

\inparagraph{Zero-shot stitching.}
Zero-shot stitching recovers dense alignment between two embedding spaces, such as vision and text.
Typically, the task assumes a sparse set of available ground-truth image-text pairs.
Stitching two unimodal pre-trained vision and language encoders is possible using relative representations \cite{Norelli:2023:ASI,Moschella:2023:RRE}, or with the help of a geometric distortion penalty \cite{Klebe:2023:GRL}.
\citet{Maiorca:2023:LST} estimate an affine transform enabling vision-text alignment of pre-trained modality-specific encoders. %
Similar to our work, \citet{Maniparambil:2024:DVL} study the similarity of the embedding spaces in the vision and text-based encoders.
However, their evaluation does not consider the fully unsupervised alignment, which is the main focus of our study.  

\inparagraph{\gw distance.}
One possibility to formulate the assignment problem between two sets is the \gw distance, which computes pairwise distances within each set.
The exact computation of the \gw distance is NP-hard; thus, its practical application has remained limited.
The few examples include shape matching \cite{Memoli:2011:GWD}, graph matching and partitioning \cite{Carin:2019:SGW,Xu:2019:GWL}, and point-cloud matching \cite{Peyre:2016:GWA}.
\citet{Alvarez-Melis:2018:GWA} apply the \gw distance for cross-lingual alignment without parallel data.
This approach realizes downstream tasks, such as word translation, in an unsupervised fashion.
\citet{Chen:2020:GOT} leverage the \gw distance for cross-modal alignment in model training.
In contrast to these works, we perform alignment between language and vision models without parallel data and any parameter training.

\inparagraph{QAP solvers.}
The quadratic assignment problem (QAP) is generally NP-hard~\cite{sahni1976p}. However, efficient approximations exist for specific problems~\cite{peyre2016gromov, hutschenreiter2021fusion}. There are many heuristics for general QAPs, which can be divided into primal and dual methods~\cite{haller2022comparative}. The primal methods try to solve an approximation of the original problem, \eg by relaxing the constraints or limiting the search space. These include the fast approximate QAP algorithm (FAQ)~\cite{vogelstein2015faq}, the 2-opt algorithm~\cite{croes1958_2opt}, and optimal transport \gw methods~\cite{peyre2016gromov}. Specific to our task, LocalCKA~\cite{Maniparambil:2024:DVL} uses the centered kernel alignment (CKA) metric and approximates the QAP with a linear assignment problem for finding vision-language correspondences.
Dual methods maximize the dual of the QAP, \eg by using block coordinate ascent~\cite{hahn1998lower} or subgradient methods~\cite{torresani2012dual}. For our heuristic, we use a dual ascent algorithm from operations research, which we refer to as \hg solver~\cite{hahn1998lower} (and detail further in \cref{sec:qap_solver}).
Finally, many methods, including ours, combine primal and dual solvers. For example, \citet{hutschenreiter2021fusion} uses fusion moves in combination with proposals from the dual formulation of the linearized QAP, thereby iteratively improving the objective. This has been shown as a useful heuristic for many computer vision problems~\cite{haller2022comparative}. 
Commercial solvers for QAPs exist, and we mainly compare with Gurobi~\cite{gurobi}, a highly optimized and general solver for mixed integer programming.

\begin{figure}
    \centering
    \includegraphics{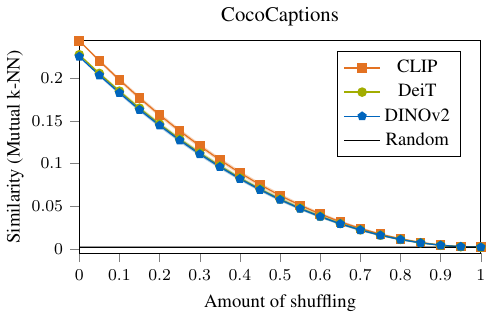}
    
    \caption{\textbf{Shuffling degrades vision-language alignment:} The vision-language alignment (here, measured by Mutual k-NN) monotonically decreases as we increasingly shuffle the oracle assignment. This holds for all considered metrics, justifying their use in the optimization objective. We encourage zooming in.}
    \label{fig:alignment_accuracy}
    \vspace{-0.5em}
\end{figure}

\section{Preliminaries}
\label{sec:prelim}
Here, we establish the link between the vision and language embeddings by formulating an alignment problem via pairwise similarities within each modality.

Let us find a correspondence between vision and language representations of $N$ classes. Given a set of images $\mathcal{I}_i$ and text descriptions $\mathcal{T}_i$, we average their embeddings from a pre-trained vision model $f_v$ and a language model $f_l$ for each class $i \in \{1, \dots, N\}$:
\begin{equation}
\label{eq:average_embeddings}
    \visionfeature_i = \frac{1}{|\mathcal{I}_i|}\sum_{I \in \mathcal{I}_i}f_v(I), \quad
    \languagefeature_i = \frac{1}{|\mathcal{T}_i|}\sum_{T \in \mathcal{T}_i}f_l(T).
\end{equation}
Moreover, we $L_2$-normalize each output of the model and the resulting averaged embeddings. We can then evaluate their similarity in each space individually using similarity measures $\visionkernel$ and $\languagekernel$.
The pairwise similarity matrices (or kernels) are then defined as \begin{equation}
    \visionpairwise_{i j} = \visionkernel(\visionfeature_i, \visionfeature_j) \: \text{and} \: \languagepairwise_{i j} = \languagekernel(\languagefeature_i, \languagefeature_j).
\end{equation}
We measure the distortion of the pairwise similarities in vision and language spaces with a distance function $l(\cdot, \cdot)$.
As its two arguments, the function takes the pairwise, possibly permuted, distance matrices from vision and language.
Summing over all distortion elements yields the distortion between the vision and language embeddings:
\begin{equation}\label{eq:distortion}
\mathcal{D}_l(\mathbf{X}, \mathbf{Y}) = \sum_{i, j= 1}^{N} l\left(\visionpairwise_{i j}, \languagepairwise_{i j}\right).
\end{equation}
Subject to a specific instantiation of $l(\cdot, \cdot)$, this formulation is general and can accommodate many existing distance (and similarity) measures.
One notable example is the \gw distance~\cite{memoli2011gromov}.
Other examples include the mutual k-nearest neighbors (Mutual k-NN)~\cite{huh2024platonic} and the centered kernel alignment (CKA)~\cite{kornblith2019similarity}.
Note that these are similarity, not distance measures.
Without loss of generality, we will use the negative inner product with these measures and build our theory around minimizing a distance measure (see \cref{sec:other_distortion_metrics}).

We now demonstrate that the pairwise distortion $\mathcal{D}_l$ is indicative of vision-language alignment.
Following \citet{Maniparambil:2024:DVL}, we begin with aligned matrices $\visionpairwise$ and $\languagepairwise$, and gradually permute the language similarity matrix $\languagepairwise$. For each permutation level, we randomly shuffle an increasing fraction of the language embeddings.
We measure the alignment in terms of Mutual k-NN and define
\begin{equation}
\label{eq:permuted_distortion}
    \mathcal{D}_\text{kNN}(\mathbf{X}, \mathbf{Y}; \pi) = \sum_{i, j= 1}^{N} l_\text{kNN}\left(\visionpairwise_{i j}, \languagepairwise_{\pi(i) \pi(j)}\right),
\end{equation}
where $\pi$ is the random permutation.
Mutual k-NN measures the average overlap between the nearest neighbors in both modalities.
In \cref{fig:alignment_accuracy}, we report the results for the CocoCaptions dataset~\cite{chen2015microsoft}, where we sample 100 random permutations for each permutation level. We observe that $\mathcal{D}_\text{kNN}$ decreases strictly monotonically as the amount of shuffling increases. This observation implies that, on average, pairwise distances between two semantically related vision and language embeddings are more similar than between two unrelated embeddings.
We confirmed this behavior for all considered datasets (\eg, ImageNet-100) and kernels (\eg, \gw distance).\footnote{The results also hold for randomly initialized vision and language networks. See \cref{sec:shuffle_vision_language} for discussion and further results.}

\begin{figure*}
\centering
\noindent\begin{minipage}[t]{0.49\textwidth}
    \centering
    \begin{algorithm}[H]
        \caption{\hg solver~\cite{hahn1998lower}}
        \label{alg:hahn_grant}
        {\small
        \begin{spacing}{1.278}
        \begin{algorithmic}[1]
           \STATE {\bfseries Input:} $\mathbf{C} \in  \R_{\geq 0}^{N \times N \times N \times N}$ cost tensor
           \STATE {\bfseries Output:} $l \leq \argmin_{\mathbf{P} \in \mathcal{P}_N} \sum_{i, j, k, l = 1}^{N} \mathbf{C}_{i j k l} \mathbf{P}_{i j} \mathbf{P}_{k l}$
           \STATE $l \leftarrow 0$
           \WHILE{not converged}
                \STATE $\texttt{leader}_{i j} \leftarrow \mathbf{C}_{i j i j}$ \hfill for $i, j \in [N]$
                \STATE \colorbox{TUMAccentOrange!20}{$\mathbf{u}, \mathbf{v}, \_ \leftarrow \text{hungarian\_matching}(\texttt{leader})$}\label{line:lin_to_const1}
                \STATE \colorbox{TUMAccentOrange!20}{$l \leftarrow l + \sum_i \mathbf{u}_i + \sum_j \mathbf{v}_j$}\label{line:lin_to_const2}
                \STATE \colorbox{TUMDarkGray!20}{$\texttt{leader}_{i j} \leftarrow \texttt{leader}_{i j} - \mathbf{u}_i - \mathbf{v}_j$}\label{line:dual1} \hfill for $i, j \in [N]$
                \STATE \colorbox{TUMYellow!20}{$\mathbf{C}_{i j k l} \leftarrow \mathbf{C}_{i j k l} + \frac{\texttt{leader}_{i j}}{N - 1}$}\label{line:redistribution} \hfill for $i \neq k, j \neq l \in [N]$
                \FOR{$i, j \in [N]$}
                    \STATE \colorbox{TUMAccentGreen!20}{$\mathbf{C}_{i j k l} \leftarrow \mathbf{C}_{i j k l} + \mathbf{C}_{k l i j}$}\hfill for $i \neq k, j \neq l \in [N]$\label{line:symmetric1}
                    \STATE \colorbox{TUMAccentGreen!20}{$\mathbf{C}_{k l i j} \leftarrow 0$}\hfill for $i \neq k, j \neq l \in [N]$\label{line:symmetric2}
                    \STATE \colorbox{TUMBlue!20}{$\mathbf{u}, \mathbf{v}, \_ \leftarrow \text{hungarian\_matching}(\mathbf{C}_{i, j, [N] \setminus \{i\}, [N] \setminus \{j\}})$}\label{line:quad_to_lin1}
                    \STATE \colorbox{TUMBlue!20}{$\mathbf{C}_{i j i j} \leftarrow \sum_k \mathbf{u}_k + \sum_l \mathbf{v}_l$}\label{line:quad_to_lin2}
                    \STATE \colorbox{TUMDarkGray!20}{$\mathbf{C}_{i j k l} \leftarrow \mathbf{C}_{i j k l} - \mathbf{u}_k - \mathbf{v}_l$}\label{line:dual2} \hfill for $i \neq k, j\neq l \in [N]$
                \ENDFOR
           \ENDWHILE
        \end{algorithmic}
        \end{spacing}
        }
    \end{algorithm}
\end{minipage}%
\hfill
\noindent\begin{minipage}[t]{0.49\textwidth}
    \centering
    \begin{algorithm}[H]
        \caption{Factorized \hg solver (Ours)}
        \label{alg:factorized_hahn_grant}
        {\small
        \begin{algorithmic}[1]
           \STATE {\bfseries Input:} $\mathbf{C}^{(1)}, \mathbf{C}^{(2)} \in  \R_{\geq 0}^{N \times N}$ symmetric cost tensors
           \STATE {\bfseries Output:} $l \leq \argmin_{\mathbf{P} \in \mathcal{P}_N} \sum_{i, j, k, l = 1}^{N} \mathbf{C}^{(1)}_{i k}\mathbf{C}^{(2)}_{j l} \mathbf{P}_{i j} \mathbf{P}_{k l}$, $\mathbf{P}^* \in \mathcal{P}_N$ permutation matrix
           \STATE $l \leftarrow 0$; $\mathbf{U}, \mathbf{V} \leftarrow \mathbf{0}_{N \times N \times N - 1}$; $\texttt{leader}_{i j} \leftarrow \mathbf{C}^{(1)}_{i i} \mathbf{C}^{(2)}_{j j}$
           \STATE \colorbox{TUMPurple!20}{$\mathbf{P}^* \leftarrow \text{primal\_heuristic}(\mathbf{C}^{(1)}, \mathbf{C}^{(2)})$}\label{line:primal_heuristic}
           \WHILE{not converged}
                \STATE $\mathbf{u}, \mathbf{v}, \mathbf{P} \leftarrow \text{lap\_solver}(\texttt{leader})$
                \STATE \colorbox{TUMPink!20}{$\mathbf{P}^* \leftarrow \text{better}(\mathbf{P}^*, \mathbf{P})$}\label{line:better1}
                \STATE $l \leftarrow l + \sum_i \mathbf{u}_i + \sum_j \mathbf{v}_j$
                \STATE $\texttt{leader}_{i j} \leftarrow \texttt{leader}_{i j} - \mathbf{u}_i - \mathbf{v}_j$ \hfill for $i, j \in [N]$
                \STATE $\mathbf{U}_{i j k} \leftarrow \mathbf{U}_{i j k} - \frac{\texttt{leader}_{i j}}{N - 1}$ \hfill for $k \neq i, j \in [N]$
                \FOR{$i, j \in [N]$}
                    \STATE $\mathbf{C}^{\text{tmp}}_{k l} \leftarrow 2\mathbf{C}^{(1)}_{i k} \mathbf{C}^{(2)}_{j l} - \mathbf{U}_{i j k} - \mathbf{V}_{i j l} - \mathbf{U}_{k l i} - \mathbf{V}_{k l j}$
                    \item[] \hfill for $i \neq k, j \neq l \in [N]$
                    \STATE $\mathbf{u}, \mathbf{v}, \mathbf{P} \leftarrow \text{lap\_solver}(\mathbf{C}^{\text{tmp}})$
                    \STATE \colorbox{TUMPink!20}{$\mathbf{P}^* \leftarrow \text{better}(\mathbf{P}^*, \mathbf{P})$}\label{line:better2}
                    \STATE $\texttt{leader}_{i j} \leftarrow \sum_k \mathbf{u}_k + \sum_l \mathbf{v}_l$
                    \STATE $\mathbf{U}_{i j k} \leftarrow \mathbf{U}_{i j k} + \mathbf{u}_k$ \hfill for $i \neq k \in [N]$
                    \STATE $\mathbf{V}_{i j l} \leftarrow \mathbf{V}_{i j l} + \mathbf{v}_l$ \hfill for $j \neq l \in [N]$
                \ENDFOR
           \ENDWHILE
        \end{algorithmic}
        }
    \end{algorithm}
\end{minipage}%
\caption{\textbf{The \hg solver (left) and the factorized \hg solver (ours, right):} The \hg solver~\cite{hahn1998lower} iteratively improves the dual bound of the QAP by solving linear assignment problems (LAPs). Our solver has higher memory efficiency for factorized cost matrices. We also use a faster solver for the LAPs and a primal heuristic that recycles the assignment from the LAPs.}
    \vspace{-0.5em}
\end{figure*}

\section{Blind matching of vision and language}
\label{sec:our_hg}
We now show that the above formulation of the vision-language alignment in terms of minimizing $\mathcal{D}_l$ leads to a quadratic assignment problem (QAP).
Then, we introduce the factorized \hg solver, which efficiently finds an approximate solution to the QAP.

\inparagraph{QAP formulation.}
The previous experiment shows that well-aligned permutations tend to have high matching accuracy.
Given unpaired vision and language data points, $\{ \visionfeature_i \}$ and $\{ \languagefeature_i \}$, we aim to find the permutation of rows and columns that minimizes their pairwise distortion, \ie
\begin{align}
    \pi^* \in &\argmin_{\pi \in \Pi^N} \sum_{i, j= 1}^{N} l\left(\visionpairwise_{i j}, \languagepairwise_{\pi(i) \pi(j)}\right),
\end{align}
where $\Pi^N$ is the permutation space, defined by bijective mappings from $\{1, \dots, N\}^N$ to $\{1, \dots, N\}^N$.
This is a quadratic assignment problem (QAP), since by exchanging the permutation space with permutation matrices, we obtain
\begin{align}
    \label{eq:qap}
    \mathbf{P}^* \in \argmin_{\mathbf{P} \in \mathcal{P}_N} \quad \sum_{i, j, k, l = 1}^{N} l\left(\visionpairwise_{i k}, \languagepairwise_{j l}\right) \mathbf{P}_{i j} \mathbf{P}_{k l}.
\end{align}
Here, $\mathcal{P}_N$ is the group of permutation matrices, \ie binary matrices in which the columns and the rows sum up to one:
\begin{equation}
\mathcal{P}_N = \{\mathbf{P} \mid \mathbf{P} \in \{0, 1\}^{N \times N}, \mathbf{P} \mathbbm{1} = \mathbbm{1}, \mathbf{P}^T \mathbbm{1} = \mathbbm{1}\}.
\end{equation}

\subsection{The \textbf{\textit{factorized}} \hg solver}
\label{sec:qap_solver}
Solving the QAP in \cref{eq:qap} by enumeration is not scalable, because there are $N!$ different permutations.
Furthermore, finding the global optimum for a general QAP is NP-hard~\cite{sahni1976p}. In practice, even proprietary solvers like Gurobi~\cite{gurobi} already fail for problem instances of size $N < 30$ (see \cref{sec:solvers}).
Nevertheless, as we discussed in \cref{sec:related}, one can recover the global solution or approximate it to a sufficient degree for some specific problem domains~\cite{vogelstein2015faq, peyre2016gromov}.
Our main technical contribution is an extension of the \hg solver that allows efficient matching of vision and language without any paired data.

\inparagraph{The \hg solver.}
The \hg solver (\cf \cref{alg:hahn_grant}) produces tight lower bounds for QAPs \cite{hahn1998lower}.
It iteratively transforms a QAP with a non-negative cost tensor $\mathbf{C}$ into an equivalent form:
\begin{align}
    &\sum_{i, j, k, l = 1}^{N} \mathbf{C}_{i j k l} \mathbf{P}_{i j} \mathbf{P}_{k l} = \\&l + \hspace{-0.4em}\sum_{i, j = 1}^{N} \text{leader}_{i j} \mathbf{P}_{i j} + \hspace{-0.9em}\sum_{i, j, k, l = 1}^{N} \hspace{-0.5em}(\mathbf{C}_{i j k l} - \mathbf{u}^{(i j)}_k - \mathbf{v}^{(i j)}_l) \mathbf{P}_{i j} \mathbf{P}_{k l}.\nonumber 
\end{align}
The solver transfers the cost from the quadratic term to a linear and then a constant term by solving linear assignment problems (LAPs). The cost of the LAPs from the quadratic term can then be moved to \texttt{leader} (\colorbox{TUMBlue!20}{Lines~\ref{line:quad_to_lin1}-\ref{line:quad_to_lin2}}), and the cost from the \texttt{leader} LAP can be moved to the constant term (\colorbox{TUMAccentOrange!20}{Lines~\ref{line:lin_to_const1}-\ref{line:lin_to_const2}}).
On the other hand, the dual solutions of the LAPs are subtracted from the cost tensor and \texttt{leader} (\colorbox{TUMDarkGray!20}{Line~\ref{line:dual1} and Line~\ref{line:dual2}}). Moreover, in each step, both $\mathbf{C}_{i j k l}$ and $\mathbf{C}_{k l i j}$ are used to improve the dual (\colorbox{TUMAccentGreen!20}{Lines~\ref{line:symmetric1}-\ref{line:symmetric2}}). At the end of each step, the non-zero values of \texttt{leader} are evenly distributed back to the corresponding submatrices in $\mathbf{C}$ (\colorbox{TUMYellow!20}{Line~\ref{line:redistribution}}). Because of the redistribution, the next iteration can use these values to make further progress. The algorithm terminates once no improvement can be made. This means that the \texttt{leader} LAP can be solved with cost zero.
Therefore, each of the LAPs corresponding to non-zero indices of the solution also has a zero-cost solution. If there is a permutation matrix that solves all of these LAPs, one has found the global optimum. In the algorithm, the constant $l$ maintains a lower bound of the original QAP.
Since we solve $N^2 + 1$ LAPs in each iteration, the algorithm has a total runtime of $\bigO(N^5)$ and a memory complexity of $\bigO(N^4)$.

\inparagraph{The \textit{factorized} \hg solver.}
We adapt the \hg solver for vision-language matching with three key extensions: \textit{(1)} We also search for primal solutions, \textit{(2)} reduce the memory complexity from $\bigO(N^4)$ to $\bigO(N^3)$, and \textit{(3)} use a faster solver for the LAPs.

We search for primal solutions in two stages. First, we initialize the problem with the FAQ~\cite{vogelstein2015faq} and 2opt~\cite{croes1958_2opt} heuristics, using the best result from 100 random seeds (\colorbox{TUMPurple!20}{Line~\ref{line:primal_heuristic}}). Second, we evaluate the solutions of each LAP on the QAP (\colorbox{TUMPink!20}{Line~\ref{line:better1} and Line~\ref{line:better2}}). This does not add considerable overhead, because the assignment was already calculated as a by-product. Therefore, we only need to evaluate the QAP objective functions once more. If the algorithm converges to a solution, that solution also needs to solve at least one of the LAPs~\cite{hahn1998lower}, and our heuristic will find the global optimum. If the algorithm does not converge, we output the lowest-cost solution found during the run.

To reduce the memory requirements, we first observe that the commonly used distortion measures $l(\cdot, \cdot)$ are decomposable as
\begin{equation}
    l(A, B) = f_1(A) + f_2(B) - h_1(A)h_2(B).
\end{equation}
This includes the squared $L_2$-distance, Kullback-Leibler divergence~\cite{peyre2016gromov}, and the negative Frobenius inner product, which we use for the centered kernel alignment (CKA) and the Mutual k-NN.
Given such a decomposable measure, the original QAP is equivalent to
\begin{align}
    \label{eq:our_qap}
    \mathbf{P}^* \in \argmin_{\mathbf{P} \in \mathcal{P}_N} \quad \sum_{i, j, k, l = 1}^{N} \mathbf{C}^{(1)}_{i k}\mathbf{C}^{(2)}_{j l} \mathbf{P}_{i j} \mathbf{P}_{k l}.
\end{align}
This is an instance of a Koopmans-Beckmann QAP~\cite{koopmans1957assignment} with cost matrices $\mathbf{C}^{(1)}=-h_1(\visionpairwise)$ and $\mathbf{C}^{(2)}=h_2(\languagepairwise)$. Although these are only two $N \times N$ matrices, applying the \hg solver would require us to compute and store the full tensor $\mathbf{C}$. Instead of updating this tensor in-place, we store them separately in $\mathbf{U}$ and $\mathbf{V}$. Therefore, the memory requirement is only $\bigO(N^3)$, instead of $\bigO(N^4)$.

Finally, we also experimented with the auction and the Jonker-Volgenant algorithm, instead of the Hungarian algorithm~\cite{kuhn1955hungarian}.
Despite the expected runtime of $\bigO(N^4 \log N)$ for the auction algorithm~\cite{bertsekas1979distributed}, we observed a faster convergence in all of our experiments using a specialized C\texttt{++} implementation~\cite{markovtsev2017lapjv} of the Jonker-Volgenant algorithm~\cite{jonker1988shortest} with a total asymptotic runtime of $\bigO(N^5)$. \cref{sec:comparison_hahn_grant_app} provides more details and a thorough comparison between our solver and the original \hg solver.

\subsection{Finding optimal matching problems}
\label{sec:optimal_subset}

Despite the strong correlation between the distortion metric and the matching accuracy (\cf \cref{fig:alignment_accuracy}), the accuracy varies significantly depending on the choice of categories.
This implies that a non-trivial matching for an arbitrary choice of semantic categories is not guaranteed.
Nevertheless, we can find a subset of matching problems where a non-trivial match is very likely, even for large problem sizes ($N > 10$).

Using the enumeration approach to find an optimal category subset is already infeasible for small problem sizes.
For example, finding a matching problem of size $N=10$ in a dataset with $100$ categories involves a search space of size $\binom{100}{10}$. However, we can explicitly formulate the problem of finding an optimal subset as an optimization problem.
For a subset of classes $S \subset \{1, \dots, L\}$, the alignment is given by
\begin{align}
    \label{eq:alignment}
    A(S) = \sum_{i, j \in S} l\left(\visionpairwise_{i j}, \languagepairwise_{i j}\right).
\end{align}
Therefore, we can find the subset $S$ that maximizes the alignment, \ie 
\begin{equation}
S^* \in \argmax_{S} A(S),
\end{equation}
where the variable $S$ satisfies $S \subset \{1, \dots, L\}$ and $|S| = N$.
Formulating this problem as a quadratic binary optimization problem~\cite{kuby1987programming}, we obtain
\begin{align}\label{eq:qop}
    \mathbf{s}^* \in &\argmax_{\mathbf{s}} \quad \sum_{i, j = 1}^{L} l\left(\visionpairwise_{i j}, \languagepairwise_{i j}\right) \mathbf{s}_{i} \mathbf{s}_{j}, \\
    &\:\text{s.t.}\quad \mathbf{s} \in \{0, 1\}^{L} \:\:\text{and}\:\: \mathbf{s}^T \mathbbm{1} = N.
\end{align}
The non-zero entries of $\mathbf{s}^*$ correspond to the classes in the subset $S^*$. This problem is an instance of the p-dispersion-sum problem~\cite{kuby1987programming}.
Although the problem is NP-hard in general~\cite{pisinger1999exact}, we can find reasonable approximations using Gurobi, as we demonstrate in \cref{sec:large_scale}.

\section{Experiments}
\label{sec:experiments}

Our experiments encompass three studies and an example application.
Our first study in \cref{sec:small_scale} focuses on a small problem size ($N = 10$), where we can analyze the properties of the global solution in a tractable manner.
We find that for an overwhelming number of pre-trained vision and language models, the optimal solutions \wrt the \gw distance correspond to non-trivial matching between the two modalities, well above the random chance.
\cref{sec:large_scale} presents the second study, where we employ our factorized \hg solver to find solutions of larger-scale problems. This study reveals that there is a subset of larger matching problems where our solver finds a non-trivial match \emph{without any paired data}.
The third study in \cref{sec:solvers} compares our factorized \hg solver to alternative optimization algorithms in terms of solution optimality. In particular, we observe that only the global optimum corresponds to non-trivial results.
Finally, \cref{sec:app} extends these studies with an exciting application, introducing unsupervised classifiers.

\inparagraph{Pre-trained models.}
In total, we consider $32$ vision models and $27$ language models using a variety of pre-trainings, architectures and sizes. For vision, we use self-supervised (DINO~\cite{Caron:2021:EPS}, DINOv2~\cite{Oquab:2024:DIN}), fully-supervised (DeiT~\cite{touvron2021training}, ConvNeXt~\cite{liu2022convnet}), and vision-language supervised (CLIP~\cite{Ramesh:2022:HTC}) models. For language, we use different models from the SentenceTransformers library~\cite{Reimers:2019:SBE} and CLIP~\cite{Ramesh:2022:HTC}. \cref{sec:models_and_datasets} elaborates further.

\inparagraph{Setup.}
Given an image classification dataset, we extract the vision embeddings for each image and compute the average vision embedding for each class. Similarly, we average the language embedding from multiple prompts to get a language representation.
To observe the effect of small perturbations in the pairwise distances on the matching, we compute the vision representation using random subsets of the data.
In our preliminary analysis, we found the \gw distance to be superior to CKA for matching vision and language.\footnote{See \cref{sec:other_distortion_metrics} for the comparison.}
Therefore, we use the \gw distance in all upcoming experiments.

\subsection{Small-scale matching}
\label{sec:small_scale}

\inparagraph{Setup.}
For small matching problems (\eg $N = 10$), we can exactly enumerate the solution space, which offers an ideal testbed for our \gw measure.
We perform matching on CIFAR-10~\cite{krizhevsky2009learning} and CINIC-10\footnote{CINIC-10 is a subset of ImageNet~\cite{russakovsky2015imagenet} with 10 classes from CIFAR-10. Our only adaption is that we keep the image resolution from ImageNet.}~\cite{darlow2018cinic}. %

\begin{figure}
    \centering
    \includegraphics{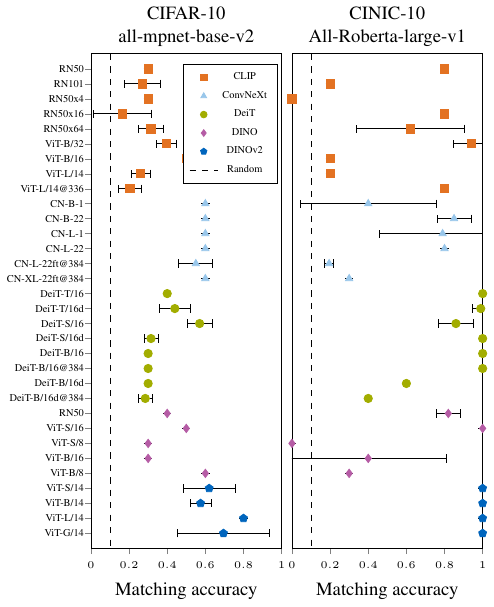}
    \vspace{-1.5em}
    \caption{\textbf{Most vision and language models can be matched non-trivially:} We visualize the accuracy for multiple vision models with the all-mpnet-base-v2~\cite{Reimers:2019:SBE} language model on CIFAR-10~\cite{krizhevsky2009learning} (left) and with the All-Roberta-large-v1~\cite{Reimers:2019:SBE} language model on CINIC-10~\cite{darlow2018cinic} (right). The error bar shows the standard deviation for 20 random seeds and the dashed line shows the performance of random matching. Observe that most vision representations can be matched with high accuracy to language, and the pre-training method has a greater impact than the model size. DINOv2~\cite{Oquab:2024:DIN} achieves the highest accuracy, on average. }
    \label{fig:matching_small_scale}
    \vspace{-0.5em}
\end{figure}

\inparagraph{Results.}
We show the matching accuracy for different vision models in \cref{fig:matching_small_scale}.
On the left, we use all-mpnet-base-v2 evaluated on CIFAR-10 and, on the right, All-Roberta-large-v1 on CINIC-10.
More combinations are given in \cref{sec:vision_language_comparison_small}.
We observe that \emph{most models perform better than 10\% accuracy}, which is the baseline for random permutations. Moreover, the DINOv2 model can be matched with the highest accuracy of 80\% on CIFAR-10 and 100\% on CINIC-10.
Curiously, we observe that it is the pre-training strategy that seems to have a larger impact than the model size. On average for all language models, the DINOv2 models are 5.3\% better on CIFAR-10 and 7.6\% better on CINIC-10 than the second-best pre-training strategy.

\subsection{Larger-scale matching}
\label{sec:large_scale}
We now investigate matching of larger problems ($N > 10$).
For vision-language matching, a larger problem implies a finer granularity of the classes (apart from the computational considerations).
This means that the pairwise distances must consistently encode more nuanced similarities. %

\inparagraph{Setup.}
We evaluate the matching accuracy on subsets of different sizes for ImageNet-100~\cite{russakovsky2015imagenet} and CIFAR-100~\cite{krizhevsky2009learning}. Following \cref{sec:optimal_subset}, we hypothesize that not all classes are represented similarly. Therefore, we use our optimization problem from \cref{sec:optimal_subset} to find subsets of classes that are well-aligned. We evaluate the alignment of the largest DINOv2, CLIP, and DeiT models with all-mpnet-base-v2.

\setlength{\tabcolsep}{2pt}
\begin{figure*}
    \centering
    \begin{subfigure}[b]{0.5\textwidth}
        \centering
        \includegraphics{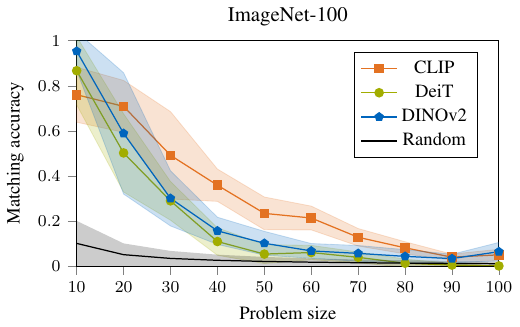}
    \end{subfigure}%
    \begin{subfigure}[b]{0.5\textwidth}
        \centering
        \includegraphics{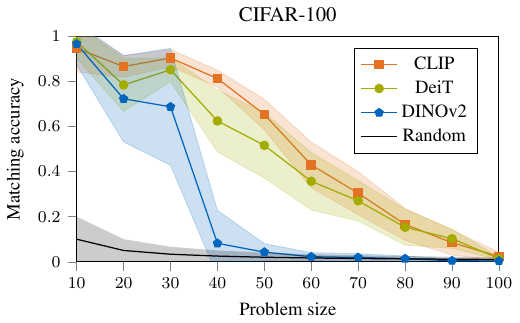}
    \end{subfigure}
    \vspace{-1.5em}
    \caption{\textbf{Some fine-grained problems can be matched with high accuracy:} For each problem size, we select the optimal ten subsets of classes using the optimization from \cref{sec:optimal_subset} on ImageNet-100~\cite{russakovsky2015imagenet} (left) and CIFAR-100~\cite{krizhevsky2009learning} (right). We show the matching accuracy for all optimization problems, each with three random seeds for the vision models DINOv2~\cite{Oquab:2024:DIN}, CLIP~\cite{Ramesh:2022:HTC}, and DeiT~\cite{touvron2021training} using the all-mpnet-base-v2 language model~\cite{Reimers:2019:SBE}. We observe that we can find optimization problems, especially for $N < 40$, that lead to accurate matching. On both datasets, CLIP performs best for most problem sizes.}
    \label{fig:matching_best_subset}
    \vspace{-0.5em}
\end{figure*}
\setlength{\tabcolsep}{6pt}

\inparagraph{Results}
\cref{fig:matching_best_subset} summarizes the results for this experiment
for ImageNet-100 (\cref{fig:matching_best_subset}, left) and for CIFAR-100 (\cref{fig:matching_best_subset}, right). We observe that \emph{all models have high accuracy for small problem sizes}, and that performance drops for larger problem sizes. Thus, for each model, there is a set of classes that is represented similarly. However, some categories adversely affect the matching, suggesting their different representation in vision and language models. As somewhat expected, CLIP outperforms the other two models in most cases, indicating that more classes share similar representation with the language model for CLIP than for DeiT and DINOv2. On CIFAR-100, the performance of DINOv2 declines more sharply than for the other models. Thus, some classes degrade the matching, if added. These results suggest that language supervision during training can align pairwise similarities, especially for fine-grained classes. Nevertheless, the self-supervised DINOv2 also learns aligned pairwise distances for some of the classes.

\subsection{Solver comparison}
\label{sec:solvers}

We now show that previous solvers are often unable to obtain meaningful assignments for our matching problems. %

\inparagraph{Setup.}
We compare the solvers both in the small-scale setting from \cref{sec:small_scale} and in the larger-scale setting using the optimal matching problems. We compare to the computer vision solver MPOpt~\cite{hutschenreiter2021fusion}, the commercial solver Gurobi~\cite{gurobi}, the fast approximate QAP algorithm (FAQ)~\cite{vogelstein2015faq}, entropic \gw optimal transport (OT)\footnote{See \cref{appendix:ot_doubly_stochastic} for how to use OT for QAPs.}~\cite{peyre2016gromov}, and LocalCKA~\cite{Maniparambil:2024:DVL}. For both experiments, we use CLIP ViT-L/14@336px and all-mpnet-base-v2.
More details and other settings are available in \cref{sec:comparison_solvers_app}.

\begin{table}[t]
    \centering
    {\small
    \begin{tabularx}{\linewidth}{
            @{}
            l
            S[table-format=3.1]
            @{$\pm$}
            S[table-format=2.1]
            S[table-format=1.2]
            @{$\pm$}
            S[table-format=1.2]
            S[table-format=3.1]
            @{$\pm$}
            S[table-format=2.1]
            @{}
        }
        \toprule
        Solver &  \multicolumn{2}{@{}c}{Accuracy (\%)} &  \multicolumn{2}{@{}c}{Cost} &  \multicolumn{2}{@{}c}{Global? (\%)} \\
        \midrule
        Random & 6.500000096857546 & 7.451598314743411 & 1.8139781839845945 & 0.20444725773313535 & 0.0 & 0.0 \\
        LocalCKA~\cite{Maniparambil:2024:DVL} & 18.500000275671482 & 29.249381708103538 & 0.5300809075686436 & 0.1601697507371285 & 5.0 & 22.3606797749979 \\
        OT~\cite{peyre2016gromov} & 33.50000027567148 & 19.808291269623112 & 1.3109703036940028 & 0.5472133884433561 & 0.0 & 0.0 \\
        FAQ~\cite{vogelstein2015faq} & 38.00000071525574 & 29.841688109602934 & 0.5461875364304327 & 0.22656515756556306 & 0.0 & 0.0 \\
        MPOpt~\cite{hutschenreiter2021fusion} & 94.00000005960464 & 18.467610154073547 & 0.3250337641736231 & 0.023686823408301753 & 90.0 & 30.779350562554622 \\
        Gurobi~\cite{gurobi} & \bfseries 100.0 & \bfseries 0.0 & \bfseries 0.31907762665707423 & \bfseries 0.01282123914296421 & \bfseries 100.0 & \bfseries 0.0 \\
        Ours & \bfseries 100.0 & \bfseries 0.0 & \bfseries 0.31907762665707423 & \bfseries 0.01282123914296421 & \bfseries 100.0 & \bfseries 0.0 \\
        \bottomrule
    \end{tabularx}
    }
    \caption{\textbf{Solver comparison on small-scale problems:} Using DINOv2~\cite{Oquab:2024:DIN} and All-Roberta-large-v1~\cite{Reimers:2019:SBE} on CIFAR-10~\cite{krizhevsky2009learning}, we report the matching accuracy, the \gw distance (cost), and the frequency of the global optimum. Most of the previous solvers fail to find the global optimum. We also observe that local optima are not sufficient to achieve a meaningful matching accuracy. By contrast, our solver always finds the global optimum, leading to the highest matching accuracy.}
    \label{tab:small_benchmark_solvers}
\end{table}

\inparagraph{Results.}
In the small-scale setting, we can compute the global optimum with enumeration. In \cref{tab:small_benchmark_solvers}, we compare the matching accuracy, the cost after optimization, and the percentage that the global optimum is reached for every method. For LocalCKA, the CKA metric is minimized using a linear approximation. We observe that minimizing this approximation does not lead to significant better results than random permutations in terms of the matching accuracy.
Both local solvers (OT, FAQ) are able to minimize the loss, but are stuck in a local minimum, thus their accuracy is often not much better than the random baseline. This shows that \emph{local optima are not sufficient to find a useful match}. MPOpt manages to approach the optimal objective, but only reaches it about 90\% of the time. In contrast, both our heuristic and Gurobi find the global optimum every time, resulting in an accuracy of 100\%.

\begin{figure}
    \centering
    \includegraphics{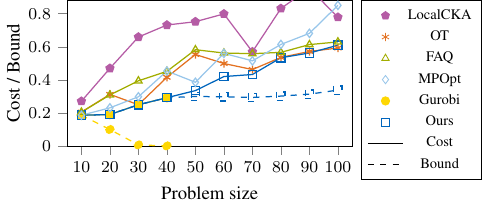}
    \vspace{-0.5em}
    \caption{\textbf{Solver comparison on larger-scale problems:} Using CLIP and all-mpnet-base-v2 on CIFAR-100~\cite{krizhevsky2009learning}, we plot the \gw distance (solid line) and its lower bound (dashed line) where available. Bounds from MPOpt~\cite{hutschenreiter2021fusion} are outside axis limits. For problems over $40$, Gurobi exceeds the $1.5$-hour time limit. Across all sizes, our solver yields tighter bounds and better primal solutions most of the time, achieving global optimality up to size $40$.}
    \label{fig:large_benchmark_solvers}
\end{figure}

\cref{fig:large_benchmark_solvers} reports the results on the larger-scale problem.
We show the cost and the lower bound for each problem size. For ease of comparison and for computational reasons, we select only one set of classes for each problem size. Consistent with the small-scale experiment, LocalCKA produces the worst values.
Interestingly, OT and FAQ are able to solve some instances with global optimality. We suspect that this is because the optimal problem may have a specialized structure. Therefore, the relaxations can lead to the global optimum for such problems. Already at size $20$, all but our method fail to reach the global optimum. For problem sizes larger than 40, Gurobi does not produce a solution within the given time limit. \emph{Our method consistently provides the tightest bounds, provably leading to a global optimum up to size $40$}. Furthermore, it achieves the best primal solution for most sizes. %

\subsection{Application: Unsupervised classifiers}
\label{sec:app}
We introduce a proof-of-concept application exploiting our formulation of the unsupervised vision-language matching. %

\inparagraph{Setup.}
Assuming no annotation, we cluster image representations using K-Means~\cite{lloyd1982least}.
We use the cluster centers for one-to-one matching with language embeddings using our QAP formulation, relying only on the pairwise distances.
Given the solution, the elements of each cluster acquire the label of the corresponding language embedding.
 
\begin{table}[t]
    \centering
    {\small
    \begin{tabularx}{\linewidth}{
            @{}
            l
            S[table-format=3.1, table-alignment=right]
            @{$\pm$}
            S[table-format=1.1, table-alignment=left]
            S[table-format=3.1, table-alignment=right]
            @{$\pm$}
            S[table-format=1.1, table-alignment=left]
            @{}
        }
        \toprule
        Model &  \multicolumn{2}{c}{All-Roberta-large-v1 (\%)} &  \multicolumn{2}{@{}c}{all-mpnet-base-v2 (\%)} \\
        \midrule
        CLIP & 28.579000085592266 & 2.306480317326015 & 28.832999914884567 & 0.7673407220116479 \\
        DeiT & 45.41200041770935 & 1.8836184400197944 & 26.82900018990039 & 3.2755070316908963 \\
        DINOv2 & \bfseries 51.14099994301796 & \bfseries 7.113629371146965 & \bfseries 37.27499991655349 & \bfseries 7.88026666341349 \\
        \bottomrule
    \end{tabularx}
    }
    \caption{\textbf{Unsupervised classification:} We combine our unsupervised matching with K-Means clustering of the image embeddings on CIFAR-10~\cite{krizhevsky2009learning}. It shows that using the cluster means as vision representation is enough to get a non-trivial accuracy. This is a good example of how unsupervised matching can enable a completely new application without supervision.}
    \label{tab:unsupervised_classification}
\end{table}

\inparagraph{Results.}
We evaluate our unsupervised classifier with several vision and language models in \cref{tab:unsupervised_classification}. The best accuracy is achieved by the DINOv2 model with 51.1\% for All-Roberta-large-v1. However, all of the considered models achieve a higher accuracy than the random baseline of 10\%. These results are clearly inferior to the supervised models.
Nevertheless, to the best of our knowledge, \emph{this is the first instance of fully unsupervised classification}. %
Furthermore, we show in \cref{sec:unsupervised_classification_solvers_app} that previous solvers are not able to find meaningful matches in this setting and analyze how clustering changes the matching performance.

\section{Discussion}
\label{sec:discussion}

\inparagraph{Can we match vision and language without paired data?}
Our small-scale experiments (\cf \cref{sec:small_scale}) indeed suggest that most vision and language models can be matched with high accuracy. This implies that on the tested datasets, the pairwise relations are similar. Furthermore, our larger-scale experiment (\cf \cref{sec:large_scale}) shows that it is also possible to match larger sets of fine-grained concepts. However, it also shows that some classes have different representation in vision and language, hence cannot be matched reliably. 

\inparagraph{Can we match arbitrary embeddings?}
Not with existing models yet.
Our larger-scale experiments in \cref{sec:large_scale} show that vision and language models may encode semantic concepts differently.
This is not surprising, since some concepts are abstract and appear only in one of the modalities, \eg ``freedom of speech''~\cite{huh2024platonic}.
For such concepts, conforming vision-language embedding spaces without explicit supervision are highly unlikely.
Another issue is symmetry: vision-language matching may be ambiguous due to multiple (equivalent) local minima.
Finally, the computational complexity scales in the order of $\bigO(N^5)$. Therefore, our solver does not scale to problems of sizes larger than those studied in \cref{sec:large_scale}. Even for problems of size $100$, the bound between the primal and the dual is significant.
Thus, reaching the global optimum becomes infeasible.

\inparagraph{Which vision model can be matched best?}
Our small-scale experiment in \cref{sec:small_scale} indicates that DINOv2 models can be matched best. However, on CIFAR-100 in the large-scale experiment, we observe that CLIP and DeiT perform better, especially on larger problems. This suggests that DINOv2 consistently represents broader concepts, whereas CLIP and DeiT better encode fine-grained relationships.

\section{Conclusion}
\label{sec:conclusion}
In this work, we investigated the ``blind'' alignment of vision and language embedding spaces, \ie without parallel data.
Our contribution combines a technical and an empirical element.
As a technical contribution, we showed that unsupervised alignment is an instance of a quadratic assignment problem, QAP.
This problem formulation includes only the pairwise similarities in each embedding space.
We proposed a new heuristic for solving the QAP, outperforming previous solvers on the alignment problem.
On the empirical side, we extensively study the alignment of pre-trained vision and language models.
Our analysis establishes feasibility of a non-trivial match between vision and language.
Although solving general and large-scale matching problems remains an open challenge, the established feasibility creates exciting avenues for novel applications.
The presented proof-of-concept unsupervised classifier is an intriguing first step in this direction.

\pagebreak
{\footnotesize \noindent\textbf{Acknowledgments.} We thank the anonymous reviewers for their valuable feedback. This work was supported by the Federal Ministry for the Environment, Nature Conservation, Nuclear Safety and Consumer Protection (BMUV) through the AuSeSol-AI project (grant 67KI21007A), and by the TUM Georg Nemetschek Institute Artificial Intelligence for the Built World (GNI) through the AICC project.
}

{
    \small
    \bibliographystyle{ieeenat_fullname}
    \bibliography{main}
}

\appendix

\clearpage
\setcounter{page}{1}
\maketitlesupplementary
\pagenumbering{roman}

\section{Overview}

\begin{itemize}
    \item \cref{sec:other_distortion_metrics} shows that our QAP formulation is general enough to accommodate the commonly used distance and similarity measures: Mutual k-NN, CKA and the \gw distance.
    \item \cref{sec:shuffle_vision_language} follows up on the experiment in \cref{sec:prelim}. It shows that the vision-language similarity measured by our distortion metrics decreases as the amount of shuffling increases. Furthermore, it elaborates on why randomly initialized networks exhibit a similar trend to pre-trained networks.
    \item \cref{sec:comparison_hahn_grant_app} compares the original \hg solver to our factorized \hg solver including a correctness proof, implementation details and an ablation showing the improvements in primal and dual estimates by our solver.
    \item \cref{sec:models_and_datasets} gives more details on the used models, datasets and specific setups for the individual experiments.
    \item \cref{sec:additional_results} provides results for other model and dataset combinations. It shows that pre-training appears more important than the model size for the matching accuracy in small-scale alignment. Also, it shows that the results for solver comparison consistently hold for all considered model and dataset combinations including unsupervised classification.
    \item \cref{appendix:ot_doubly_stochastic} shows that the \gw optimal transport solvers can be seen as solvers for a relaxation of our QAP.
\end{itemize}

\section{\!Mutual k-NN and CKA as distortion metrics}
\label{sec:other_distortion_metrics}
In \cref{sec:our_hg}, we introduce the notation of distortion metrics in terms of a unimodal kernel function $\visionkernel$ and $\languagekernel$, and a distance function $l$, which is decomposable as
\begin{equation}
\label{eq:l_decomposition_app}
    l(A, B) = f_1(A) + f_2(B) - h_1(A)h_2(B).
\end{equation}
In this section, we show how Mutual k-NN~\cite{huh2024platonic}, the centered kernel alignment (CKA)~\cite{kornblith2019similarity}, and the \gw (\gwshort) distance~\cite{memoli2011gromov} fit into this formulation.

\inparagraph{Mutual k-NN~\cite{huh2024platonic}.}
Mutual k-NN is defined as the average overlap between the nearest neighbors in both modalities, \ie, for one sample, it is
\begin{align}
    m_{\text{kNN}}(\visionfeature_i, \languagefeature_i) = \frac{1}{k}|\topk_k^{\visionfeature}(i) \cap \topk_k^{\visionfeature}(i)|.
\end{align}
Here, we define the top-\textit{k} indices in terms of the highest inner product as $\topk_k^{\visionfeature}(i)$. Following \citet{huh2024platonic}, we exclude $i$ from this set.
To include this into our framework, we define the kernel function as 
\begin{align}
    \visionkernel^{\text{kNN}}(\visionfeature_i, \visionfeature_j) = \frac{1}{\sqrt{N k}} \mathbbm{1}[j \in \topk_k^{\visionfeature}(i)],
\end{align}
and accordingly for the language embeddings. Then, the similarity matrices are
\begin{align}
    \visionpairwise_{i j}^{\text{kNN}} = \visionkernel^{\text{kNN}}(\visionfeature_i, \visionfeature_j) \: \text{and} \: \languagepairwise_{i j}^{\text{kNN}} = \languagekernel^{\text{kNN}}(\languagefeature_i, \languagefeature_j).
\end{align}
Furthermore, we use the negative inner product as our distortion function:
\begin{align}
    l_{\text{inner}}(A, B) = - A \cdot B,
\end{align}
which trivially satisfies \cref{eq:l_decomposition_app} with $f_1 = f_2 = 0$ and $h_1$, $h_2$ being identity functions.
This choice of kernels and the distance metric leads to the Mutual k-NN metric in our framework:
\begin{equation}
\begin{aligned}
    \mathcal{D}&_\text{kNN}(\visionpairwise^{\text{kNN}}, \languagepairwise^{\text{kNN}})\\
    &= \sum_{i, j= 1}^{N} l\left(\visionpairwise_{i j}^{\text{kNN}}, \languagepairwise_{i j}^{\text{kNN}}\right)\\
    &= -\sum_{i, j= 1}^{N} \visionkernel^{\text{kNN}}(\visionfeature_i, \visionfeature_j) \languagekernel^{\text{kNN}}(\languagefeature_i, \languagefeature_j) \\
    &= -\sum_{i, j= 1}^{N} \frac{1}{\sqrt{N k}} \mathbbm{1}[j \in \topk_k^{\visionfeature}(i)] \frac{1}{\sqrt{N k}} \mathbbm{1}[j \in \topk_k^{\languagefeature}(i)] \\
    &= -\frac{1}{N}\sum_{i, j= 1}^{N} \frac{1}{k} \mathbbm{1}[j \in \topk_k^{\visionfeature}(i) \land j \in \topk_k^{\languagefeature}(i)] \\
    &= -\frac{1}{N}\sum_{i, j= 1}^{N} \frac{1}{k} \mathbbm{1}[j \in \topk_k^{\visionfeature}(i) \cap \topk_k^{\languagefeature}(i)] \\
    &= -\frac{1}{N}\sum_{i = 1}^{N} \frac{1}{k} |j \in \topk_k^{\visionfeature}(i) \cap \topk_k^{\languagefeature}(i)| \\
    &= -\frac{1}{N}\sum_{i = 1}^{N} m_{\text{kNN}}(\visionfeature_i, \languagefeature_i).
\end{aligned}    
\end{equation}

\inparagraph{CKA.}
We derive CKA~\cite{kornblith2019similarity, Maniparambil:2024:DVL} in a similar fashion. For a kernel function $\hat{k}$ and kernel matrices $\hat{\visionpairwise}_{i j} = \hat{k}(\visionfeature_i, \visionfeature_j)$ and $\hat{\languagepairwise}_{i j} = \hat{k}(\languagefeature_i, \languagefeature_j)$, the CKA is defined as
\begin{align}
    \text{CKA}(\hat{\visionpairwise}, \hat{\languagepairwise}) = \frac{\tr(\hat{\visionpairwise} \mathbf{C} \hat{\languagepairwise} \mathbf{C})}{\sqrt{\tr(\hat{\visionpairwise} \mathbf{C} \hat{\visionpairwise} \mathbf{C}) \tr(\hat{\languagepairwise} \mathbf{C} \hat{\languagepairwise} \mathbf{C})}},
\end{align}
where $\mathbf{C} = \mathbf{I} - \frac{1}{N} \mathbbm{1} \mathbbm{1}^T$. Similar to previous work~\cite{Maniparambil:2024:DVL}, we use the linear kernel in this work. Our kernel matrices are
\begin{equation}
    \begin{aligned}
        &\visionpairwise^{\text{CKA}} = \frac{\hat{\visionpairwise} \mathbf{C}}{\sqrt{\tr(\hat{\visionpairwise} \mathbf{C} \hat{\visionpairwise} \mathbf{C})}} \,\, \quad \text{and} \\
        &\languagepairwise^{\text{CKA}} = \frac{\mathbf{C}^T \hat{\languagepairwise}^T}{\sqrt{\tr(\hat{\languagepairwise} \mathbf{C} \hat{\languagepairwise} \mathbf{C})}}.
    \end{aligned}
\end{equation}

\begin{table}[t]
    \centering
    {\small
    \begin{tabularx}{\linewidth}{
            @{}
            X
            S[table-format=2.1, table-alignment=right]
            @{$\pm$}
            S[table-format=2.1, table-alignment=left]
            @{\hspace{5em}}
            S[table-format=2.1, table-alignment=right]
            @{$\pm$}
            S[table-format=1.1, table-alignment=left]
            @{}
        }
        \toprule
        {Models} & \multicolumn{2}{l}{CIFAR-10 (\%)} & \multicolumn{2}{@{}l}{CINIC-10 (\%)} \\
        \midrule
        Mutual k-NN & 59.00000214576722 & 3.077935790092365 & 70.0 & 0.0 \\
        CKA & 63.50000202655792 & 7.451597622732691 & \bfseries 80.0 & \bfseries 0.0 \\
        \gwshort Distance & \bfseries 69.50000047683716 & \bfseries 24.165003315452655 & 79.00000095367432 & 3.0779357900923654 \\
        \bottomrule
    \end{tabularx}
    }
    \caption{\textbf{\gw distance is the best measure for matching:} We show the accuracy for CIFAR-10~\cite{krizhevsky2009learning} and CINIC-10~\cite{darlow2018cinic} using DINOv2~\cite{Oquab:2024:DIN} and all-mpnet-base-v2~\cite{Reimers:2019:SBE} using Mutual k-nearest neighbor (Mutual k-NN)~\cite{huh2024platonic}, centered kernel alignment (CKA)~\cite{kornblith2019similarity}, and the \gw (\gwshort) distance~\cite{memoli2011gromov} as a metric. The \gwshort distance leads to the good matching accuracies for both datasets.}
    \label{tab:cifar10_cinic10_accuracy_kernels}
    \vspace{-0.5em}
\end{table}

\noindent Using the negative inner product as the distance metric leads to the negative CKA:
\begin{align}
    \mathcal{D}&_\text{CKA}(\visionpairwise^{\text{CKA}}, \languagepairwise^{\text{CKA}})\\
    &= \sum_{i, j= 1}^{N} l\left(\visionpairwise_{i j}^{\text{CKA}}, \languagepairwise_{i j}^{\text{CKA}}\right)\nonumber\\
    &= - \sum_{i, j= 1}^{N} \frac{(\hat{\visionpairwise} \mathbf{C})_{i j}}{\sqrt{\tr(\hat{\visionpairwise} \mathbf{C} \hat{\visionpairwise} \mathbf{C})}} \frac{(\mathbf{C}^T \hat{\languagepairwise}^T)_{i j}}{\sqrt{\tr(\hat{\languagepairwise} \mathbf{C} \hat{\languagepairwise}}}\nonumber\\
    &= - \frac{1}{\sqrt{\tr(\hat{\visionpairwise} \mathbf{C} \hat{\visionpairwise} \mathbf{C}) \tr(\hat{\languagepairwise} \mathbf{C} \hat{\languagepairwise} \mathbf{C})}} \sum_{i, j= 1}^{N} (\hat{\visionpairwise} \mathbf{C})_{i j}(\hat{\languagepairwise} \mathbf{C})_{j i} \nonumber\\
    &= - \frac{1}{\sqrt{\tr(\hat{\visionpairwise} \mathbf{C} \hat{\visionpairwise} \mathbf{C}) \tr(\hat{\languagepairwise} \mathbf{C} \hat{\languagepairwise} \mathbf{C})}} \tr(\hat{\visionpairwise} \mathbf{C}\hat{\languagepairwise} \mathbf{C}) \nonumber\\
    &= - \text{CKA}(\hat{\visionpairwise}, \hat{\languagepairwise}) \nonumber
\end{align}

\inparagraph{\gwshort distance~\cite{memoli2011gromov}.}
For the \gwshort distance, we can choose the $L_2$-norm as the kernel and the squared distance as the distortion metric:
\begin{equation}
\begin{aligned}
    \visionkernel^{\text{\gwshort}}(\visionfeature_i, \visionfeature_j) &= \|\visionfeature_i - \visionfeature_j\|_2, \\
    \languagekernel^{\text{\gwshort}}(\languagefeature_i, \languagefeature_j) &= \|\languagefeature_i - \languagefeature_j\|_2, \\
    l_{\text{\gwshort}}(A, B) &= (A - B)^2.
\end{aligned}
\end{equation}
Then, the \gwshort distance is given by
\begin{align}
    \mathcal{D}&_\text{\gwshort}(\visionpairwise^{\text{\gwshort}}, \languagepairwise^{\text{\gwshort}}) = \sum_{i, j= 1}^{N} \left(\|\visionfeature_i - \visionfeature_j\|_2 - \|\languagefeature_i - \languagefeature_j\|_2\right)^2.
\end{align}
The objective function after finding the optimal permutation matrix similar to \cref{eq:qap} is then equivalent to the original \gwshort distance comparing two metric-measure spaces~\cite{peyre2016gromov}.

\cref{tab:cifar10_cinic10_accuracy_kernels} empirically compares these formulations in a small-scale data regime, using CIFAR-10~\cite{krizhevsky2009learning} and CINIC-10~\cite{darlow2018cinic} datasets introduced in the main text. %
We find that despite the wider adoption of Mutual k-NN and CKA metrics in previous work, the \gw distance leads to a comparable or higher matching accuracy on both datasets.

\section{Shuffled vision-language alignment}
\label{sec:shuffle_vision_language}

In this section, we elaborate on the setup for the shuffling experiment from \cref{sec:prelim}.
As we claimed in the main text (\cf \cref{sec:prelim}), the observations are consistent accross all tested datasets and models, which we also report here.

\inparagraph{Setup.}
Given aligned image and language representations, $(\visionfeature_i, \languagefeature_i)$, and a shuffling level $\alpha \in [0, 1]$, we randomly choose $\floor*{\alpha N}$ elements that are randomly permuted. Every other element is kept in place. Afterwards, the distortion of this permutation is computed with \cref{eq:distortion} or \cref{eq:permuted_distortion} after the permutation. Note that we use the image embeddings here instead of the averaged object embeddings. For classification datasets, we take the same language embedding for all elements of that class. However, we have seen in preliminary experiments that the curve looks similar when considering the averaged object representations instead of the image representations. We plot $21$ equidistant shuffling levels at $\alpha \in \{0, 0.05, \dots, 1\}$. Each level is based on $100$ random seeds to sample the subset and permutation.

\begin{figure*}
    \centering
    \includegraphics{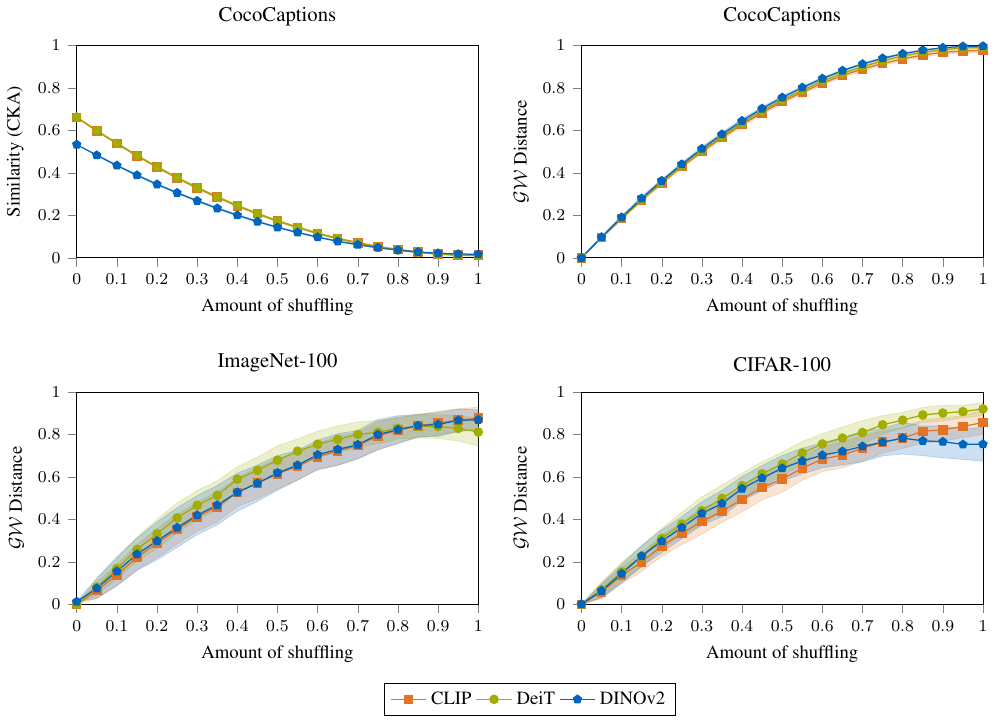}
    
    \caption{\textbf{Shuffling degrades vision-language alignment:} We report CKA similarity and the \gw distance on three datasets and based on five encoding methods, including pre-trained and randomly initialized networks, and raw pixel values. Similar to our observation with Mutual k-NN, the CKA and \gw distance correlate strongly \wrt the amount of shuffling, which suggests that these metrics are also suitable measures for blind vision-language matching.}
    \label{fig:alignment_accuracy_app}
    \vspace{-0.8em}
\end{figure*}

\inparagraph{Shuffling with other kernels, datasets, and models.}
In addition to \cref{fig:alignment_accuracy}, which uses the CocoCaptions dataset~\cite{chen2015microsoft}, and Mutual k-NN as the distortion metric, we show more combinations in \cref{fig:alignment_accuracy_app}. In this setting, Mutual k-NN is only meaningful for paired datasets because the k-nearest neighbors are ambiguous when language features are replicated. For the \gwshort distance, the pairwise distances are also dependent on the dimensionality of the embedding spaces. Therefore, we standardize the distance to be in the range between zero and one for this shuffling experiment. We observe that the similarity (/ distortion) decreases (/ increases) strongly monotonically with more shuffling. This behavior is consistent for all considered datasets and metrics. This observation suggests that the pairwise relations are more similar between the semantically corresponding vision and language representations than between the non-semantic ones.

\begin{figure*}
    \centering
    \includegraphics{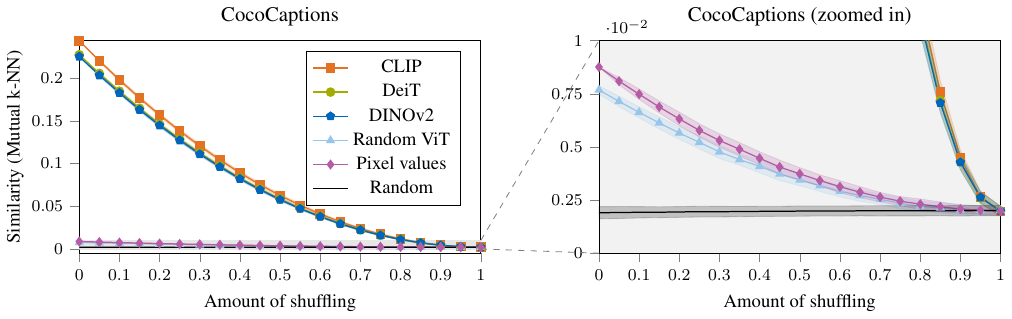}
    
    \caption{\textbf{Randomly initialized networks behave similarly to the pre-trained ones:} Zooming in (right) on the randomly initialized ViT, we observe a strikingly similar qualitative behavior -- the similarity decreases monotonically with the increased degree of shuffling. This observation also holds for the curve resulting from raw pixel values. This surprising phenomenon presumably originates from the properties of the natural image manifold and the Lipschitz-continuity of neural networks.}
    \label{fig:alignment_accuracy_random}
    \vspace{-0.8em}
\end{figure*}

\inparagraph{A note on randomly initialized networks.}
In \cref{fig:alignment_accuracy_random}, we present a plot of the experiment in \cref{fig:alignment_accuracy} with the addition of randomly initialized ViT-H/14~\cite{Dosovitskiy:2021:AIW} models, which were initiated with $20$ distinct random seeds.
Additionally, we demonstrate the distortion of a representation based on the stacked pixel values, \ie, without any neural network.
Furthermore, we show a zoomed-in version on the left-hand side to illustrate the behavior on a finer scale.

\begin{figure}
    \centering
    \includegraphics{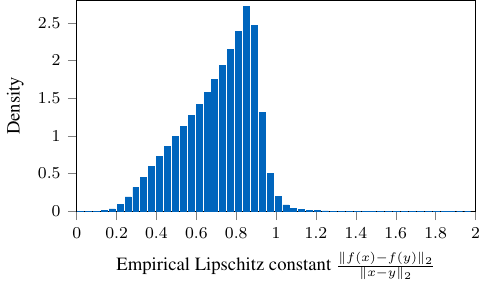}
    
    \caption{\textbf{Random networks roughly preserve distances:} Observe that the mode of the empirical Lipschitz constant is close to $1$, which suggests that network encoding approximately preserves the distances in the input domain.}
    \label{fig:lipschitz_density}
    \vspace{-0.8em}
\end{figure}

We observe that the alignment of randomly initialized networks exhibits the same monotonically decreasing behavior with increased amount of shuffling, as for pre-trained networks (despite the absolute alignment value being smaller).
For images, this can be explained by the similarity of the pixel distribution within each semantic category (\eg, a green landscape for animal stock). These similarities appear more frequently for semantically affiliated classes and can dominate the pairwise distance encoding.
To understand the behavior for randomly initialized networks, we plot the distribution of the \emph{Empirical Lipschitz constant} in \cref{fig:lipschitz_density}, defined by
\begin{equation}
    \mathcal{K}_\text{Lipschitz}(\mathbf{x}, \mathbf{y}) = \frac{\|f(\mathbf{x}) - f(\mathbf{y})\|_2}{\|\mathbf{x} - \mathbf{y}\|_2},
\end{equation}
for samples $\mathbf{x}$, $\mathbf{y}$ and function $f$. 
Intuitively, it measures the degree of distance distortion in the output space \wrt the input space for each sample pair in the dataset.
Here, we use CocoCaptions and the same language model as in \cref{fig:alignment_accuracy_random}. We observe that most values are close to one. This implies that the distance after encoding remains approximately preserved. 
Nevertheless, the distances are slightly distorted, which explains why the absolute similarity in \cref{fig:alignment_accuracy_random} is lower for random ViTs than for the pixel values. 
In summary, shuffling reduces alignment even for the distance in terms of pixel values. This then transfers to random ViTs because these distances are approximately preserved.

\begin{figure*}
\centering
\setcounter{algorithm}{0}
\noindent\begin{minipage}[t]{0.49\textwidth}
    \centering
    \begin{algorithm}[H]
        \caption{\hg solver~\cite{hahn1998lower}}
        \label{alg:hahn_grant_app}
        {\small
        \begin{spacing}{1.3187}
        \begin{algorithmic}[1]
           \STATE {\bfseries Input:} $\mathbf{C} \in  \R_{\geq 0}^{N \times N \times N \times N}$ cost tensor
           \STATE {\bfseries Output:} $l \leq \argmin_{\mathbf{P} \in \mathcal{P}_N} \sum_{i, j, k, l = 1}^{N} \mathbf{C}_{i j k l} \mathbf{P}_{i j} \mathbf{P}_{k l}$
           \STATE $l \leftarrow 0$
           \WHILE{not converged}
                \STATE $\texttt{leader}_{i j} \leftarrow \mathbf{C}_{i j i j}$ \hfill for $i, j \in [N]$\label{line:leader_update_app}
                \STATE $\mathbf{u}, \mathbf{v}, \_ \leftarrow \text{hungarian\_matching}(\texttt{leader})$\label{line:lin_to_const1_app}
                \STATE $l \leftarrow l + \sum_i \mathbf{u}_i + \sum_j \mathbf{v}_j$\label{line:lin_to_const2_app}
                \STATE $\texttt{leader}_{i j} \leftarrow \texttt{leader}_{i j} - \mathbf{u}_i - \mathbf{v}_j$\label{line:dual1_app} \hfill for $i, j \in [N]$
                \STATE $\mathbf{C}_{i j k l} \leftarrow \mathbf{C}_{i j k l} + \frac{\texttt{leader}_{i j}}{N - 1}$\label{line:redistribution_app} \hfill for $i \neq k, j \neq l \in [N]$
                \FOR{$i, j \in [N]$}
                    \STATE $\mathbf{C}_{i j k l} \leftarrow \mathbf{C}_{i j k l} + \mathbf{C}_{k l i j}$\hfill for $i \neq k, j \neq l \in [N]$\label{line:symmetric1_app}
                    \STATE $\mathbf{C}_{k l i j} \leftarrow 0$\hfill for $i \neq k, j \neq l \in [N]$\label{line:symmetric2_app}
                    \STATE $\mathbf{u}, \mathbf{v}, \_ \leftarrow \text{hungarian\_matching}(\mathbf{C}_{i, j, [N] \setminus \{i\}, [N] \setminus \{j\}})$
                    \STATE $\mathbf{C}_{i j i j} \leftarrow \sum_k \mathbf{u}_k + \sum_l \mathbf{v}_l$\label{line:quad_to_lin2_app}
                    \STATE $\mathbf{C}_{i j k l} \leftarrow \mathbf{C}_{i j k l} - \mathbf{u}_k - \mathbf{v}_l$\label{line:dual2_app} \hfill for $i \neq k, j\neq l \in [N]$
                \ENDFOR
           \ENDWHILE
        \end{algorithmic}
        \end{spacing}
        }
    \end{algorithm}
\end{minipage}%
\hfill
\noindent\begin{minipage}[t]{0.49\textwidth}
    \centering
    \begin{algorithm}[H]
        \caption{Factorized \hg solver (Ours)}
        \label{alg:factorized_hahn_grant_app}
        {\small
        \begin{algorithmic}[1]
           \STATE {\bfseries Input:} $\mathbf{C}^{(1)}, \mathbf{C}^{(2)} \in  \R_{\geq 0}^{N \times N}$ symmetric cost tensors
           \STATE {\bfseries Output:} $l \leq \argmin_{\mathbf{P} \in \mathcal{P}_N} \sum_{i, j, k, l = 1}^{N} \mathbf{C}^{(1)}_{i k}\mathbf{C}^{(2)}_{j l} \mathbf{P}_{i j} \mathbf{P}_{k l}$, $\mathbf{P}^* \in \mathcal{P}_N$ permutation matrix
           \STATE $l \leftarrow 0$; $\mathbf{U}, \mathbf{V} \leftarrow \mathbf{0}_{N \times N \times N - 1}$; $\texttt{leader}_{i j} \leftarrow \mathbf{C}^{(1)}_{i i} \mathbf{C}^{(2)}_{j j}$
           \STATE $\mathbf{P}^* \leftarrow \text{primal\_heuristic}(\mathbf{C}^{(1)}, \mathbf{C}^{(2)})$\label{line:primal_heuristic_app}
           \WHILE{not converged}\label{line:while_app}
                \STATE $\mathbf{u}, \mathbf{v}, \mathbf{P} \leftarrow \text{lap\_solver}(\texttt{leader})$\label{line:lap_solver_app}
                \STATE $\mathbf{P}^* \leftarrow \text{better}(\mathbf{P}^*, \mathbf{P})$\label{line:better1_app}
                \STATE $l \leftarrow l + \sum_i \mathbf{u}_i + \sum_j \mathbf{v}_j$
                \STATE $\texttt{leader}_{i j} \leftarrow \texttt{leader}_{i j} - \mathbf{u}_i - \mathbf{v}_j$ \hfill for $i, j \in [N]$\label{line:dual1_app2}
                \STATE $\mathbf{U}_{i j k} \leftarrow \mathbf{U}_{i j k} - \frac{\texttt{leader}_{i j}}{N - 1}$ \hfill for $k \neq i, j \in [N]$\label{line:redistribution_app2}
                \FOR{$i, j \in [N]$}
                    \STATE $\mathbf{C}^{\text{tmp}}_{k l} \leftarrow 2\mathbf{C}^{(1)}_{i k} \mathbf{C}^{(2)}_{j l} - \mathbf{U}_{i j k} - \mathbf{V}_{i j l} - \mathbf{U}_{k l i} - \mathbf{V}_{k l j}$\label{line:tmp_app}
                    \item[] \hfill for $i \neq k, j \neq l \in [N]$
                    \STATE $\mathbf{u}, \mathbf{v}, \mathbf{P} \leftarrow \text{lap\_solver}(\mathbf{C}^{\text{tmp}})$\label{line:lap_solver_app2}
                    \STATE $\mathbf{P}^* \leftarrow \text{better}(\mathbf{P}^*, \mathbf{P})$\label{line:better2_app}
                    \STATE $\texttt{leader}_{i j} \leftarrow \sum_k \mathbf{u}_k + \sum_l \mathbf{v}_l$
                    \STATE $\mathbf{U}_{i j k} \leftarrow \mathbf{U}_{i j k} + \mathbf{u}_k$ \hfill for $i \neq k \in [N]$\label{line:dual2_app2}
                    \STATE $\mathbf{V}_{i j l} \leftarrow \mathbf{V}_{i j l} + \mathbf{v}_l$ \hfill for $j \neq l \in [N]$\label{line:dual2_app3}
                \ENDFOR
           \ENDWHILE
        \end{algorithmic}
        }
    \end{algorithm}
\end{minipage}%
\caption{\textbf{The \hg solver (left) and the factorized \hg solver (ours, right):} The \hg solver~\cite{hahn1998lower} iteratively improves the dual bound of the QAP by solving linear assignment problems (LAPs). Our solver improves the memory requirements of the \hg solver for factorized cost matrices, introduces a primal heuristic that reuses the assignment from the LAPs, and uses a faster solver for the LAPs.}
    \vspace{-0.5em}
\end{figure*}

\section{On the \emph{factorized} \hg solver}
\label{sec:comparison_hahn_grant_app}
In \cref{sec:our_hg}, we recap the \hg solver~\cite{hahn1998lower} and introduce our \emph{factorized} \hg solver. Here, we provide more details on the implementation of the algorithms and show that both algorithms result in the same lower bounds. We also include an empirical study supporting the design choices behind our factorized \hg solver: the factorization, faster LAP solvers, and finding primal solutions.

\subsection{Proof of equivalence to \hg solver}
The main idea of the proof is that the lower bound, the \texttt{leader}, and all LAPs are the same for both algorithms. The main difference is that the dual vectors are stored in the tensors $\mathbf{U}$ and $\mathbf{V}$ instead of updating $\mathbf{C}$ in-place.

First, we assume that $\mathbf{C}_{ijkl} = \mathbf{C}^{(1)}_{ik} \mathbf{C}^{(2)}_{jl}$ with $\mathbf{C}^{(1)}$ and $\mathbf{C}^{(2)}$ being non-negative and symmetric. Given this assumption, we will show that \cref{alg:hahn_grant_app} and \cref{alg:factorized_hahn_grant_app} are equivalent without our adaptions (Lines~\ref{line:primal_heuristic_app}, \ref{line:better1_app}, and \ref{line:better2_app}) and using the Hungarian matching as the LAP solver. We will show that both algorithms have the same lower bound $l$, the \texttt{leader}, and the sum of complementary costs $\mathbf{C}_{i j k l} + \mathbf{C}_{k l i j}$. In particular, denoting the variables from \cref{alg:factorized_hahn_grant_app} with a bar, we have
\begin{align}
\label{eq:qed}
    &l = \overline{l}, \quad \texttt{leader} = \overline{\texttt{leader}}, \quad \text{and}\\
    &\mathbf{C}_{i j k l} + \mathbf{C}_{k l i j} = 2\overline{\mathbf{C}}^{(1)}_{i k} \overline{\mathbf{C}}^{(2)}_{j l} - \overline{\mathbf{U}}_{i j k} - \overline{\mathbf{V}}_{i j l} - \overline{\mathbf{U}}_{k l i} - \overline{\mathbf{V}}_{k l j}.\nonumber
\end{align}
for all $i \neq k, j \neq l \in [N]$.
\noindent Starting with the first iteration until after \cref{alg:hahn_grant_app}, Line~\ref{line:leader_update_app} and \cref{alg:factorized_hahn_grant_app}, Line~\ref{line:while_app}, we have that the lower bound $l = \overline{l} = 0$ and 
\begin{align}
    \texttt{leader} = \mathbf{C}_{i j i j} = \mathbf{C}^{(1)}_{ii} \mathbf{C}^{(2)}_{jj} = \overline{\texttt{leader}}
\end{align}
are initialized the same. Furthermore, because $\overline{\mathbf{U}} = \overline{\mathbf{V}} = 0$, we have that
\begin{align}
    \mathbf{C}_{i j k l} = \mathbf{C}^{(1)}_{ik} \mathbf{C}^{(2)}_{jl} - \overline{\mathbf{U}}_{i j k} - \overline{\mathbf{V}}_{i j l}.
\end{align}
Thus, both algorithms have an equivalent starting condition. 

Next, given that \cref{eq:qed} holds, we show that both values are changed in an equivalent way. Given the same $\texttt{leader}$ matrix, Line~\ref{line:lin_to_const1_app}-\ref{line:dual1_app} from \cref{alg:hahn_grant_app} change $l$ and $\texttt{leader}$ in the same way as Lines~\ref{line:lap_solver_app}-\ref{line:dual1_app2} from \cref{alg:factorized_hahn_grant_app}. Moreover, after \cref{alg:hahn_grant_app}, Line~\ref{line:redistribution_app} and \cref{alg:factorized_hahn_grant_app}, Line~\ref{line:redistribution_app2}, $\mathbf{C}_{i j k l}$ and $\mathbf{U}_{i j k}$ are changed in the same way. Therefore, the sum is also preserved:
\begin{equation}
\begin{aligned}
    \mathbf{C}&_{i j k l} + \mathbf{C}_{k l i j}\\
    = &\mathbf{C}^{\text{prev}}_{i j k l} + \mathbf{C}^{\text{prev}}_{k l i j} + \frac{\texttt{leader}_{i j}}{N - 1} + \frac{\texttt{leader}_{k l}}{N - 1}\\
    = &2\overline{\mathbf{C}}^{(1)}_{i k} \overline{\mathbf{C}}^{(2)}_{j l} - \overline{\mathbf{U}}^{\text{prev}}_{i j k} + \frac{\texttt{leader}_{i j}}{N - 1} - \overline{\mathbf{V}}_{i j l}\\ 
    &- \overline{\mathbf{U}}^{\text{prev}}_{k l i} + \frac{\texttt{leader}_{k l}}{N - 1} - \overline{\mathbf{V}}_{k l j} \\
    = &2\overline{\mathbf{C}}^{(1)}_{i k} \overline{\mathbf{C}}^{(2)}_{j l} - \overline{\mathbf{U}}_{i j k} - \overline{\mathbf{V}}_{i j l} - \overline{\mathbf{U}}_{k l i} - \overline{\mathbf{V}}_{k l j},
\end{aligned}
\end{equation}
where the superscipt $(^\text{prev})$ denotes the value before executing the line.

In Line~\ref{line:symmetric1_app} and Line~\ref{line:symmetric2_app} of \cref{alg:hahn_grant_app}, the values from the complementary positions are redistributed to the current submatrix. Line~\ref{line:tmp_app} from \cref{alg:factorized_hahn_grant_app} also aggregates the sum of the complementary elements, albeit not by changing the cost matrices. Therefore, the sum of both elements remains the same. By definition of $\mathbf{C}^{\text{tmp}}$, it follows, that $\mathbf{C}_{i j k l} = \mathbf{C}^{\text{tmp}}_{k l}$ and the Hungarian matching produces the same values for both algorithms. As the next step, the objective value is added to the $\texttt{leader}$ in \cref{alg:hahn_grant_app} by first adding it to $\mathbf{C}_{i j i j}$ in Line~\ref{line:quad_to_lin2_app} and then to $\texttt{leader}_{i j}$ in Line~\ref{line:leader_update_app} in the next iteration. In \cref{alg:factorized_hahn_grant_app}, this value is directly added to $\texttt{leader}_{i j}$. Finally, the dual variables are subtracted from the cost in Line~\ref{line:dual2_app} of \cref{alg:hahn_grant_app} and Line~\ref{line:dual2_app2} and Line~\ref{line:dual2_app3} of \cref{alg:factorized_hahn_grant_app}. This preserves the sum:
\begin{equation}
\begin{aligned}
    \mathbf{C}_{i j k l} &+ \mathbf{C}_{k l i j} = \\
    &= \mathbf{C}^{\text{prev}}_{i j k l}  - \mathbf{u}_k - \mathbf{v}_l + \mathbf{C}^{\text{prev}}_{k l i j} - \mathbf{u}_i - \mathbf{v}_j\\
    &= 2\overline{\mathbf{C}}^{(1)}_{i k} \overline{\mathbf{C}}^{(2)}_{j l} - \overline{\mathbf{U}}^{\text{prev}}_{i j k} - \mathbf{u}_k - \overline{\mathbf{V}}^{\text{prev}}_{i j l} - \mathbf{v}_l \\
    &\quad - \overline{\mathbf{U}}^{\text{prev}}_{k l i} - \mathbf{u}_i - \overline{\mathbf{V}}^{\text{prev}}_{k l j} - \mathbf{v}_j\\
    &= 2\overline{\mathbf{C}}^{(1)}_{i k} \overline{\mathbf{C}}^{(2)}_{j l} - \overline{\mathbf{U}}_{i j k} - \overline{\mathbf{V}}_{i j l} - \overline{\mathbf{U}}_{k l i} - \overline{\mathbf{V}}_{k l j}.
\end{aligned}
\end{equation}
Therefore, each iteration in both algorithms changes the costs in an equivalent way. As a result, the final bound $l$ and each solution to the LAPs are the same. $\square$ %

In practice, we can also remove the non-negativity constraint because adding a constant to $\mathbf{C}^{(1)}$ or $\mathbf{C}^{(2)}$ leads to an equivalent optimization problem with an additional constant term, \ie
\begin{align}
\label{eq:const_add}
    \sum_{i, j, k, l = 1}^{N} (\mathbf{C}^{(1)}_{i k} &+ c)\mathbf{C}^{(2)}_{j l} \mathbf{P}_{i j} \mathbf{P}_{k l} = \\
    &= \sum_{i, j, k, l = 1}^{N} \mathbf{C}^{(1)}_{i k}\mathbf{C}^{(2)}_{j l} \mathbf{P}_{i j} \mathbf{P}_{k l} + c \sum_{j, l = 1}^{N} \mathbf{C}^{(2)}_{j l}.\nonumber
\end{align}
Therefore, we can subtract the minimal element from both matrices, apply the algorithm to the resulting non-negative matrices, and subtract the constant from \cref{eq:const_add} to the final objective value to retrieve the optimal objective value of the original problem.

\subsection{Implementation details}
We implement both \cref{alg:factorized_hahn_grant_app} and \cref{alg:hahn_grant_app} in Python using PyTorch~\cite{pytorch}. The main computational bottleneck is the LAP solver. Therefore, we use a custom C\texttt{++} algorithm for the forward-reverse auction~\cite{forwardreverse} algorithm and for the Jonker-Volgenant algorithm~\cite{jonker1988shortest, markovtsev2017lapjv}. We stop the algorithm if the relative or absolute improvement of the dual bound $l$ is smaller than $\varepsilon=1e-6$ or the primal objective is within $\varepsilon$ of the dual bound. Finally, for the auction algorithm, a larger relaxation $\varepsilon^{\text{auc}}$ usually leads to a faster runtime with the drawback of worse objectives and dual vectors. The \hg algorithm also works for suboptimal dual vectors, but we observe that towards the end of the algorithm, better solutions in the LAPs are required. Therefore, we initialize $\varepsilon^{\text{auc}} = 0.1$ relatively high in the beginning and multiply it with a factor of $0.9$ in every iteration. %

\begin{table}
    \centering
    {\small
    \sisetup{exponent-mode=scientific, round-mode=figures, round-precision=2, exponent-base=e, exponent-product=}
    \begin{tabularx}{\linewidth}{
            @{}
            l
            @{\hspace{0.9em}}
            S[table-format=1.6, table-alignment=right]
            @{\hspace{2pt}}
            S[table-format=1.6, table-alignment=left]
            @{\hspace{0.9em}}
            S[table-format=1.6, table-alignment=right]
            @{\hspace{2pt}}
            S[table-format=1.6, table-alignment=left]
            @{}
        }
        \toprule
        Solver & \multicolumn{2}{c}{Cost} & \multicolumn{2}{c}{Bound} \\
        \midrule
        \hg & -1.979372 &  & {--} &  \\
        ~+ factorized & -1.979380 & \qsdec{\num{-7.929687499963478e-06}} & -2.097599 & \qlinc{} \\
        ~+ auction & -1.979379 & \qsinc{\num{9.374999996314415e-07}} & -1.980904 & \qlinc{\num{0.11669480468750004}} \\
        ~+ Jonker-Volgenant & -1.979391 & \qsdec{\num{-1.1874999999772484e-05}} & -1.979865 & \qlinc{\num{0.0010389453125001502}} \\
        ~+ LAP solutions & -1.979430 & \qsdec{\num{-3.835937500018538e-05}} & -1.979865 & \qeq{} \\
        ~+ primal heuristic & -1.979521 & \qsdec{\num{-9.171874999958085e-05}} & -1.979866 & \qldec{\num{-1.2109375000513012e-06}} \\
        \bottomrule
    \end{tabularx}
    }
    \caption{\textbf{Cost and bounds for our Hahn-Grant adaptation ($N=100$):} The factorization and the auction algorithm slightly increase the cost of the solution, while leading to a small bound. The Jonker-Volgenant and especially the LAP solutions and primal heuristics also lead to better primal solutions.}
    \label{tab:ablation_hahn_grant100}
    \vspace{-0.5em}
\end{table}

\subsection{Ablations}
\inparagraph{Setup.}
We compare the original \hg solver with our adaptation from \cref{sec:our_hg}: the factorization into matrices $\mathbf{C}^{(1)}$ and $\mathbf{C}^{(2)}$, using the auction algorithm or the Jonker-Volgenant as faster LAP solvers (Line~\ref{line:lap_solver_app} and Line~\ref{line:lap_solver_app2}), evaluating the individual LAP primal solutions (Line~\ref{line:better1_app} and Line~\ref{line:better2_app}), and using primal heuristics (Line~\ref{line:primal_heuristic_app}). For these experiments, we generate two sets with $100$ vectors of dimensionality $1024$ each, drawn element-wise from a standard normal distribution. These vectors are normalized, and the pairwise inner product is computed to produce two $100 \times 100$ similarity matrices. We apply all variations of the algorithms with a time limit of two hours to these cost matrices and evaluate the quality of the primal solutions and of the bound. We repeat the experiment with $5$ random seeds and average the results. We also repeat the experiment with a set of $40$ random vectors and a time limit of one hour.

\inparagraph{Results.}
\cref{tab:ablation_hahn_grant100} shows the result for each adaptation for size $100$ and \cref{tab:ablation_hahn_grant40} for size $40$. We observe that the original \hg solver reaches a non-trivial primal-dual bound for size $40$ but does not finish the first iteration within the two hour time limit for size $100$. Even though factorization was introduced for improved memory, it also slightly speeds up the computation, leading to a smaller bound for both sizes. The auction and Jonker-Volgenant algorithms were introduced to improve the speed of the algorithm and, therefore, lead to smaller bounds. 
However, the Jonker-Volgenant algorithm leads to the tightest bounds empirically for large problems while the auction algorithm is slightly better on the size $40$ problem. The LAP solutions do not change the dual estimate but improve the primal solution by a significant margin. Finally, the primal heuristics further improve the primal solution with the trade-off of slightly worse bounds. Since we are mostly interested in good primal solutions and measuring the quality of the solution, we use the algorithm with all the adaptations. %

\begin{table}
    \centering
    {\small
    \sisetup{exponent-mode=scientific, round-mode=figures, round-precision=2, exponent-base=e, exponent-product=}
    \begin{tabularx}{\linewidth}{
            @{}
            l
            @{\hspace{0.9em}}
            S[table-format=1.6, table-alignment=right]
            @{\hspace{2pt}}
            S[table-format=1.6, table-alignment=left]
            @{\hspace{0.9em}}
            S[table-format=1.6, table-alignment=right]
            @{\hspace{2pt}}
            S[table-format=1.6, table-alignment=left]
            @{}
        }
        \toprule
        Solver & \multicolumn{2}{c}{Cost} & \multicolumn{2}{c}{Bound} \\
        \midrule
        \hg & -1.949088 & & -2.056868 & \\
        ~+ factorized & -1.949109 & \qsdec{\num{-2.0904541015553946e-05}} & -1.950544 & \qlinc{\num{0.10632415771484394}} \\
        ~+ auction & -1.949133 & \qsdec{\num{-2.429199218756395e-05}} & -1.949529 & \qlinc{\num{0.0010142517089841707}} \\
        ~+ Jonker-Volgenant & -1.949111 & \qsinc{\num{2.1972656250035527e-05}} & -1.949577 & \qldec{\num{-4.7882080077776834e-05}} \\
        ~+ LAP solutions & -1.949211 & \qsdec{\num{-0.00010003662109392408}} & -1.949577 & \qeq{} \\
        ~+ primal heuristic & -1.949294 & \qsdec{\num{-8.245849609345157e-05}} & -1.949577 & \qeq{} \\
        \bottomrule
    \end{tabularx}
    }
    \caption{\textbf{Cost and bounds for our Hahn-Grant adaptation ($N=40$):} Our adaptations exhibit strong benefits either in terms of the solution cost or the tightness of the bound, frequently both. The auction algorithm leads to tighter bounds than the Jonker-Volgenant algorithm.}
    \label{tab:ablation_hahn_grant40}
\end{table}

\section{Experimental details}
\label{sec:models_and_datasets}
In this section, we provide more details supplementing the experiments in \cref{sec:experiments}, which were excluded from the main text due to space constraints.

\begin{figure*}
    \centering
    \includegraphics{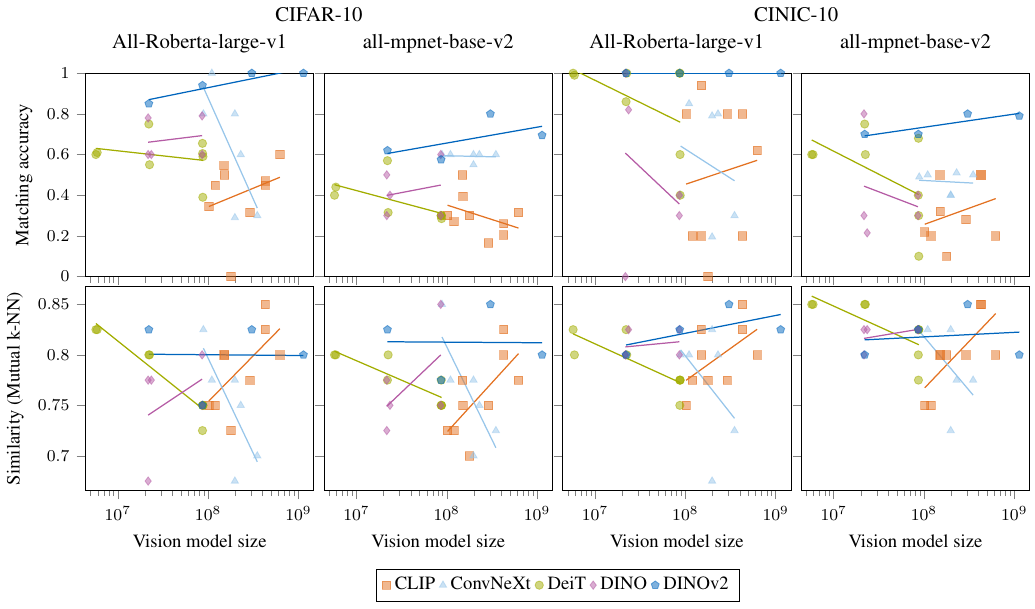}
    \caption{\textbf{Matching accuracy and Mutual k-NN on CIFAR-10 and CINIC-10:} Larger models do not necessarily lead to stronger alignment with language. Instead, it is the pre-training method that characterizes the vision-language alignment most. Here, DINOv2 yields the highest matching accuracy for both datasets and language models.}
    \label{fig:size_vs_class_small}
\end{figure*}

Our method only assumes pre-computed embeddings (or already pre-computed similarity matrices). Therefore, we are not limited to specific architectures or pre-training strategies. Our goal is to choose a variety of different architectures, pre-trainings, and model sizes for both modalities.

\inparagraph{Vision models.}
We consider self-supervised, fully-supervised and vision-language supervised models with convolutional and vision transformer architectures of different capacities. For self-supervised methods, we consider different models from DINO~\cite{Caron:2021:EPS} trained on ImageNet-1k~\cite{russakovsky2015imagenet} and models from DINOv2~\cite{Oquab:2024:DIN} trained on LVD-142M. For supervised models, DeiT~\cite{touvron2021training} and ConvNeXt~\cite{liu2022convnet} pre-trained on ImageNet-1k, ImageNet-22k~\cite{russakovsky2015imagenet} and a a combination of both are used. Furthermore, we choose CLIP~\cite{Ramesh:2022:HTC} as our vision-language model with both ResNet and ViT backbones. We use the code and models from the official repository except for ConvNeXt, where we use pre-trained models from the \texttt{timm} library. In total, we use $32$ vision models.

\inparagraph{Language models.}
Similar to the vision models, we consider different pre-trainings and network sizes. In particular, in addition the the CLIP~\cite{Ramesh:2022:HTC} text models, contrastive learning, question-answer models, and average word embeddings are considered. We use the official CLIP repository for all CLIP text models and the SentenceTransformers~\cite{Reimers:2019:SBE} library for all other models. In total, we use $27$ language models.

\inparagraph{Datasets.} 
We evaluate our experiments on CIFAR-10~\cite{krizhevsky2009learning}, CINIC-10~\cite{darlow2018cinic}, CIFAR-100~\cite{krizhevsky2009learning}, and ImageNet-100~\cite{russakovsky2015imagenet}. For ImageNet-100, we use the validation split and for all other datasets, we use the test split. We choose the same class names and prompts as ASIF~\cite{Norelli:2023:ASI} whenever available, and follow the same preprocessing as CLIP otherwise. For ImageNet-100, we observed an improved performance by encoding each class as ``$<$name$>$: $<$definition$>$`` using the WordNet definitions and more descriptive class names.

\inparagraph{General setup.}
We implement all experiments in Python with PyTorch~\cite{pytorch}.
In general, we pre-compute the embeddings for each vision and language model and every dataset using PyTorch Lightning~\cite{pytorchlightning}.
We do not use GPUs except for precomputing the embeddings.
After computing the image-wise embeddings, we normalize them and average them for every class, see \cref{eq:average_embeddings}. The resulting object-wise embedding is again normalized. To evaluate the impact of small changes in the embeddings (and pairwise similarities), we only take a random subset of the image representations to compute the average. In particular, we use a half of the embeddings drawn uniformly for every random seed. For the language embeddings, we take the average over the embeddings for all different prompts.

\inparagraph{Small-scale matching.}
We use $20$ random seeds and all models on CIFAR-10 and CINIC-10 for small-scale matching. In each experiment, we enumerate all permutations and compute all costs explicitly, returning the permutation with the minimal cost. The comparison of all combinations of vision and language models is given in \cref{sec:vision_language_comparison_small}.

\inparagraph{Larger-scale matching.}
For larger-scale matching, we use different subsets of classes using the optimization problem introduced in \cref{sec:optimal_subset} solved with Gurobi~\cite{gurobi}. For each problem size $N \in \{10, 20, \dots, 100\}$, we use the $10$ best subset of classes, each with $3$ random seeds for computing the object-level vision embeddings. We use our factorized \hg solver and a time limit of one hour for each matching problem. The models are DINOv2 ViT-G/14, CLIP ViT-L/14@336px, and the distilled DeiT-B/16@384px with all-mpnet-base-v2.

\inparagraph{Solver comparison.}
We evaluate the solvers both on small-scale and larger-scale matching. For small-scale matching, we use $20$ random seeds for DINOv2 ViT-G/14, CLIP ViT-L/14@336px, and the distilled DeiT-B/16@384px with all-mpnet-base-v2 and All-Roberta-large-v1 on CIFAR-10 and CINIC-10. We present the results for all combinations in \ref{sec:comparison_solvers_app}. The larger-scale benchmark follows the setting of larger-scale matching, but only considers one out of the ten subsets and one random seed for CIFAR-100 using CLIP CLIP ViT-L/14@336px and all-mpnet-base-v2 with a time limit of 1.5 hours. 

\inparagraph{Unsupervised classifier.}
For unsupervised classification, we use K-Means~\cite{lloyd1982least} to cluster image embeddings into prototype (object) embeddings. We use K-Means\texttt{++}~\cite{kmeanspp} from Scikit-learn~\cite{scikit-learn} with $100$ initializations. The cluster centers are then matched to the language embeddings using our factorized \hg solver. Similar to the previous experiments, we only use a random 50\% subset of the vision embeddings and evaluate the method for $20$ random seeds. We report the results for DINOv2 ViT-G/14, CLIP ViT-L/14@336px, and the distilled DeiT-B/16@384px with all-mpnet-base-v2 and All-Roberta-large-v1 on CIFAR-10.

\begin{table}[t]
    \centering
    {\small
    \begin{tabularx}{\linewidth}{
            @{}
            l
            @{\hspace{1.5em}}
            S[table-format=2.1, table-alignment=right]
            @{$\pm$}
            S[table-format=2.1, table-alignment=left]
            @{\hspace{2.5em}}
            S[table-format=6.1, table-alignment=right]
            @{$\pm$}
            S[table-format=2.1, table-alignment=left]
            @{}
        }
        \toprule
        Solver & \multicolumn{2}{@{}l}{Accuracy (\%)} & \multicolumn{2}{c}{Matching Accuracy (\%)}\\
        \midrule
        Random & 10.399000009056177 & 9.712256939027188 & 11.000000201165674 & 12.096106618426884 \\
        LocalCKA~\cite{Maniparambil:2024:DVL} & 13.009000080637628 & 11.315274232818341 & 13.000000305473803 & 14.179303282269135 \\
        OT~\cite{peyre2016gromov} & 3.694999990984793 & 4.178284055744621 & 3.5000000521540633 & 4.893604922216324 \\
        FAQ~\cite{vogelstein2015faq} & 10.669000120833513 & 13.85568724212689 & 10.500000268220901 & 15.719582685803868 \\
        MPOpt~\cite{hutschenreiter2021fusion} & 20.63300031004473 & 22.4647538286765 & 22.5 & 25.520889276702025 \\
        Ours & \bfseries 45.41200041770935 & \bfseries 1.8836184400197944 & \bfseries 50.50000011920929 & \bfseries 2.2360685106199445 \\
        \color{TUMGray}GT & \color{TUMGray}84.64000046253204 & \color{TUMGray}0.41584439539033224 & \color{TUMGray}100 & \color{TUMGray}0 \\
        \bottomrule
    \end{tabularx}
    }
    \caption{\textbf{Solver comparison on unsupervised classification:} Using DeiT and All-Roberta-large-v1, we show the accuracy of the unsupervised classifier for different solvers. The right column further shows how many of the centroids are mapped to the best class for the given cluster in line with the matching accuracy from the unsupervised matching experiments. Our QAP solver achieves a considerably higher matching and classification accuracy than the other methods.}
    \label{tab:unsupervised_classification_solvers}
\end{table}

\section{Evaluation results}
\label{sec:additional_results}
Following up on \cref{sec:experiments}, we report the results for all vision and language models in the small-scale matching setting (\cf \cref{sec:vision_language_comparison_small}) and benchmark of the different solvers using multiple datasets and models (\cf \cref{sec:unsupervised_classification_solvers_app} and \cref{sec:comparison_solvers_app}).

\subsection{Comparison of vision and language models}
\label{sec:vision_language_comparison_small}

In this section, we report the results on the small-scale matching, spanning all vision and language models in our study. The results show, in particular, that DINOv2 outperforms the other models on both datasets and that the model size is less important than the pre-training method.

\begin{figure*}
    \centering
    \includegraphics{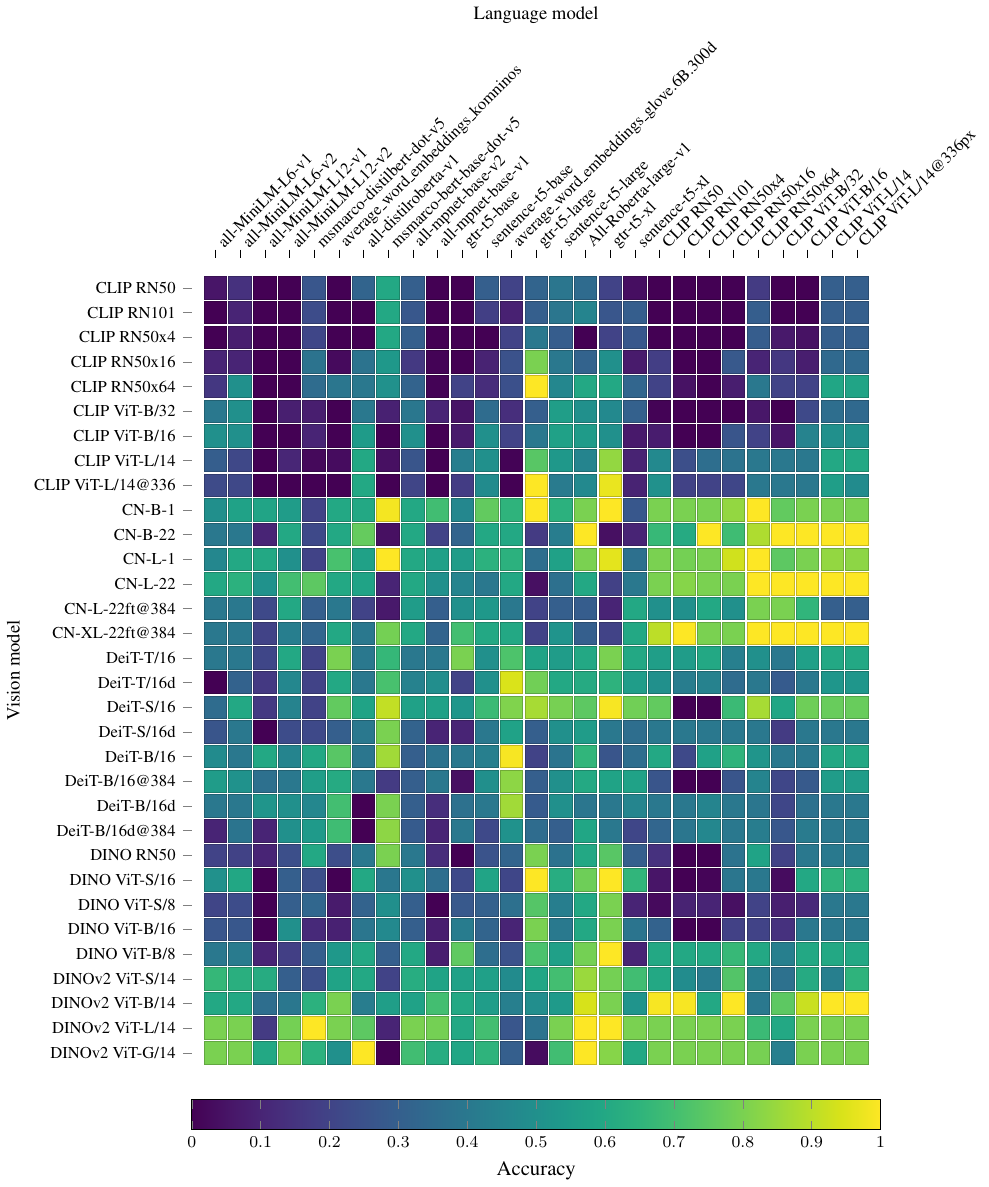}
    \caption{\textbf{Vision-language alignment accuracy on CIFAR-10:} We observe that the pre-training strategy tends to have a more impactful role on vision-language alignment than the model capacity. Here, DINOv2 and ConvNeXt (\eg CN-B-1) families exhibits a prominent degree of image-text alignment.}
    \label{fig:heatmap_cifar10}
\end{figure*}

\begin{figure*}
    \centering
    \includegraphics{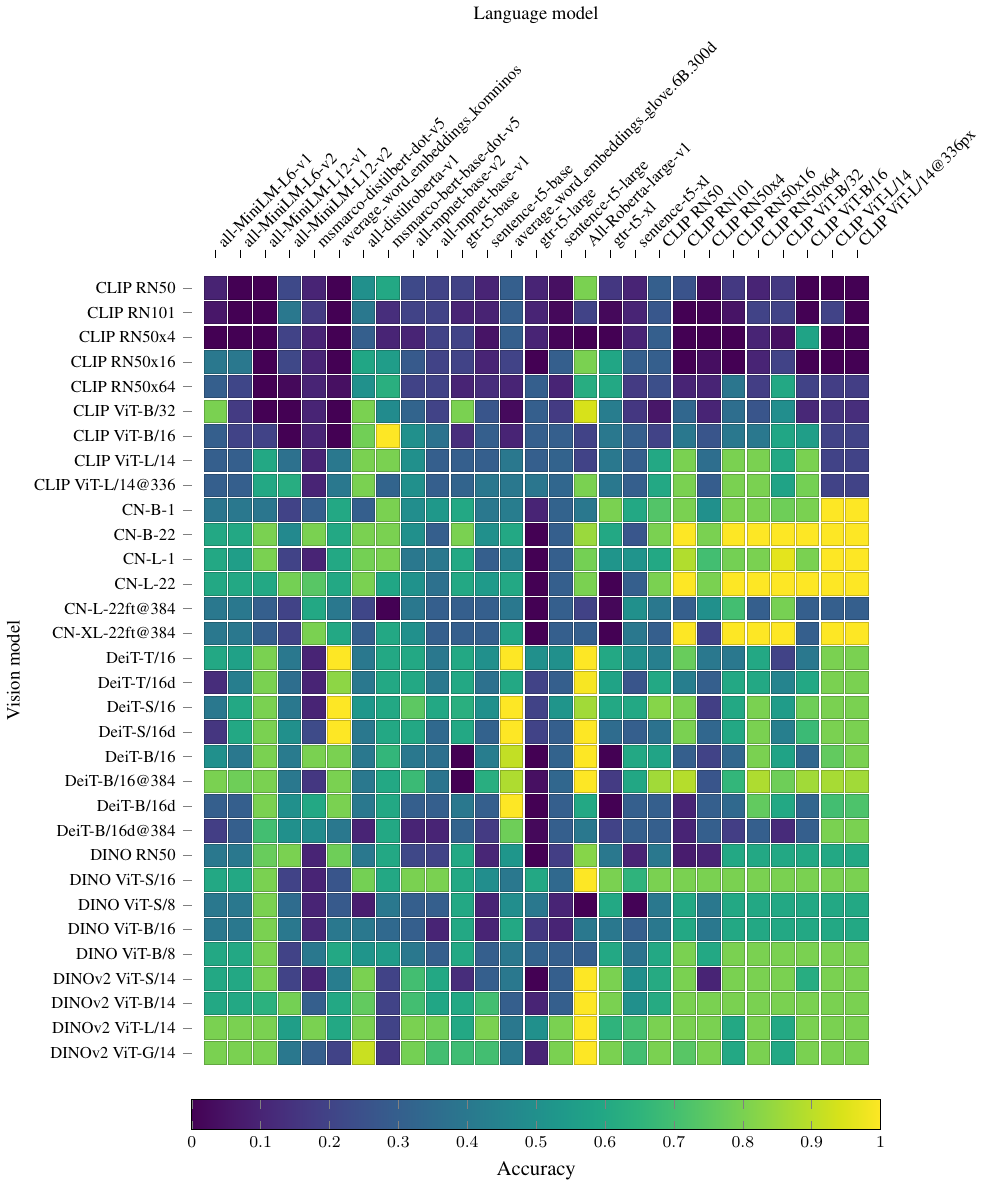}
    \caption{\textbf{Vision-language alignment accuracy on CINIC-10:} Here, DINO-based pre-training in conjunction with the CLIP-based text embedding exhibit a higher degree of conformity.}
    \label{fig:heatmap_cinic10}
\end{figure*}

In \cref{fig:size_vs_class_small}, we show the matching accuracy (top) and Mutual k-NN with $k = 5$ (bottom) of each vision model in combination with All-Roberta-large-v1 (left) and all-mpnet-base-v2 (right) for CIFAR-10 (left) and CINIC-10 (right). The lines show the trend for increasing model sizes that are fitted to the models of varying capacity for each model class (different colors). The model size corresponds to the number of parameters. First, we observe that for both datasets and language models, DINOv2 yields the highest matching accuracy. Furthermore, for every model class, there is no a clear propensity of larger models to perform better. As this seems to contradict the platonic representation hypothesis \cite{huh2024platonic}, we report the Mutual k-NN in the bottom plots.
We observe that in line with our conclusions, there is still no clear improvement of the Mutual k-NN metric \wrt an increasing model capacity. This implies that larger models do not necessarily lead to stronger alignment with language -- at least on CIFAR-10 and CINIC-10. Considering  the observations by \citet{huh2024platonic}, this suggests that scaling the models could lead to a better alignment on the Wit dataset~\cite{wit}, even though it may not be sufficient to improve the alignment on every other dataset. We report the individual matching accuracy of each combination of vision and language model for CIFAR-10 in \cref{fig:heatmap_cifar10} and for CINIC-10 in \cref{fig:heatmap_cinic10}.


\subsection{Unsupervised classification: solver comparison}
\label{sec:unsupervised_classification_solvers_app}
We also compare the solvers on the unsupervised classification setting in \cref{tab:unsupervised_classification_solvers}. 
In addition to the classification accuracy, we also report the matching accuracy.
Given a clustering and the ground truth labels, we compute the optimal ground truth matching. 
The matching accuracy evaluates how well our unsupervised matching coincides with the ground truth matching.
Finally, we report the performance when the ground truth matching is used instead of our unsupervised matching to evaluate the quality of the clustering.

Similar to the unsupervised matching experiments, we observe that our solver outperforms the other solvers in terms of both classification accuracy and matching accuracy. 
This means that our solver finds matches that agree well with the ground truth matches.
We also observe that the k-means clustering is not perfect, and approximately $15\%$ of the images are clustered in the wrong clusters.
This hurts our clustering in two ways.
First, since only whole clusters are assigned to classes, misclassified images will remain in the respective (wrong) cluster. 
Second, incorrect samples distort the centroids, which in turn can further affect the pairwise distances, leading to a worse matching.
However, even with imperfect centroids, our matching finds non-trivial matches and leads to the first non-trivial unsupervised classification, which was not possible using previous local solvers.

\subsection{Comparison of solvers on small-scale problems}
\label{sec:comparison_solvers_app}
Similar to \cref{tab:small_benchmark_solvers}, we benchmark different solvers for all combinations in \cref{tab:small_benchmark_solvers_app}. Our factorized \hg solver and Gurobi~\cite{gurobi} find the global optimum for all problems. For most combinations, these solvers are also the only solvers finding the global optimum. MPOpt~\cite{hutschenreiter2021fusion} also finds the global optimum for some problems (\eg, CINIC-10 with DINOv2) but fails for other problems (\eg, CIFAR-10 with CLIP). For all but four experiments (all-mpnet-base-v2 on CIFAR-10 for all three vision models and on CINIC-10 with DeiT), the global optimum also leads to the best matching accuracy, which underlines the importance of the \gw distance as a proxy measure for blind matching. Finally, the matching accuracy is close to random for most solvers (LocalCKA~\cite{Maniparambil:2024:DVL}, OT~\cite{peyre2016gromov}, and FAQ~\cite{vogelstein2015faq}).
This shows that a global solver is crucial to obtain non-trivial results.

\begin{table*}[p]
    \centering
    {\small
    \begin{tabularx}{\linewidth}{
            @{}
            l
            l
            l
            @{\hspace{1.5em}}
            S[table-format=3.1]
            @{$\pm$}
            S[table-format=2.1]
            S[table-format=1.2]
            @{$\pm$}
            S[table-format=1.2]
            S[table-format=3.1]
            @{$\pm$}
            S[table-format=2.1]
            S[table-format=3.1]
            @{$\pm$}
            S[table-format=2.1]
            S[table-format=1.2]
            @{$\pm$}
            S[table-format=1.2]
            S[table-format=3.1]
            @{$\pm$}
            S[table-format=2.1]
            @{}
        }
        \toprule
         &  &  & \multicolumn{6}{c}{all-mpnet-base-v2} & \multicolumn{6}{c}{All-Roberta-large-v1} \\
         &  &  & \multicolumn{2}{c}{Accuracy (\%)} & \multicolumn{2}{c}{Cost} & \multicolumn{2}{c}{Global? (\%)} & \multicolumn{2}{c}{Accuracy (\%)} & \multicolumn{2}{c}{Cost} & \multicolumn{2}{c}{Global? (\%)} \\
        \midrule
        \multirow[c]{21}{*}{\rotatebox[origin=c]{90}{CIFAR-10}} & \multirow[t]{7}{*}{CLIP} & Random & 6.500000096857546 & 7.451598314743411 & 1.881639268153982 & 0.17832089252987846 & 0.0 & 0.0 & 6.500000096857546 & 7.451598314743411 & 1.8396716908825035 & 0.16444210860185696 & 0.0 & 0.0 \\
        &  & LocalCKA~\cite{Maniparambil:2024:DVL} & \bfseries 25.000000335276123 & \bfseries 10.513149724166185 & 0.3532610756982228 & 0.04551671282200578 & 60.0 & 50.262468995003466 & 17.000000588595864 & 15.927467845464419 & 0.5876591860505043 & 0.12186809983162686 & 0.0 & 0.0 \\
        &  & OT~\cite{peyre2016gromov} & 0.0 & 0.0 & 0.3525607172787413 & 0.005656402448167416 & 5.0 & 22.360679774997898 & 10.00000014901161 & 0.0 & 0.6352606422562006 & 0.01565473922569881 & 0.0 & 0.0 \\
        &  & FAQ~\cite{vogelstein2015faq} & 12.00000017881393 & 10.05249394879452 & 0.3455753615278625 & 0.01708857138462059 & 50.0 & 51.29891760425771 & 0.5000000074505805 & 2.236068010819799 & 0.7912100248154452 & 0.045692774859772724 & 0.0 & 0.0 \\
        &  & MPOpt~\cite{hutschenreiter2021fusion} & 0.0 & 0.0 & 0.61007835131508 & 0.009040267331939885 & 0.0 & 0.0 & 0.0 & 0.0 & 0.5200354029117957 & 0.009470084247150175 & 0.0 & 0.0 \\
        &  & Gurobi~\cite{gurobi} & 20.50000041723251 & 6.048053463200611 & \bfseries 0.33435985672606355 & \bfseries 0.007591666357085671 & \bfseries 100.0 & \bfseries 0.0 & \bfseries 47.000000178813934 & \bfseries 7.326950533930157 & \bfseries 0.4003024779764151 & \bfseries 0.006951030535930551 & \bfseries 100.0 & \bfseries 0.0 \\
        &  & Ours & 20.50000041723251 & 6.048053463200611 & \bfseries 0.33435985672606355 & \bfseries 0.007591666357085671 & \bfseries 100.0 & \bfseries 0.0 & \bfseries 47.000000178813934 & \bfseries 7.326950533930157 & \bfseries 0.4003024779764151 & \bfseries 0.006951030535930551 & \bfseries 100.0 & \bfseries 0.0 \\
        \cline{2-15}
        & \multirow[t]{7}{*}{DeiT} & Random & 6.500000096857546 & 7.451598314743411 & 1.8273495878421866 & 0.191619531395305 & 0.0 & 0.0 & 6.500000096857546 & 7.451598314743411 & 1.8170277316816221 & 0.21593638246292668 & 0.0 & 0.0 \\
        &  & LocalCKA~\cite{Maniparambil:2024:DVL} & 24.000000879168503 & 9.947229662463615 & 0.33655444260796197 & 0.0376854340591296 & 0.0 & 0.0 & 20.000000372529026 & 5.619515092789624 & 1.7277943008333068 & 0.5387716309363021 & 0.0 & 0.0 \\
        &  & OT~\cite{peyre2016gromov} & 12.000000178813933 & 4.103913469493691 & 0.3200519687163165 & 0.03515824902205031 & 5.0 & 22.360679774997898 & 10.00000014901161 & 0.0 & 1.4267747357704432 & 0.010009596719478035 & 0.0 & 0.0 \\
        &  & FAQ~\cite{vogelstein2015faq} & \bfseries 40.00000026077032 & \bfseries 15.217717993753002 & 0.2960918057664002 & 0.014441817435391267 & 0.0 & 0.0 & 22.500000335276123 & 9.665456813609138 & 0.3305212658073608 & 0.01840652457427534 & 0.0 & 0.0 \\
        &  & MPOpt~\cite{hutschenreiter2021fusion} & 0.0 & 0.0 & 0.7213535037555552 & 0.010376510584860926 & 0.0 & 0.0 & 0.0 & 0.0 & 0.746604158290648 & 0.02017690689325576 & 0.0 & 0.0 \\
        &  & Gurobi~\cite{gurobi} & 28.500001057982438 & 3.6634758128654634 & \bfseries 0.27077751850336074 & \bfseries 0.003795094831243368 & \bfseries 100.0 & \bfseries 0.0 & \bfseries 59.00000214576721 & \bfseries 3.0779357900923654 & \bfseries 0.27330570541418614 & \bfseries 0.004909022473389803 & \bfseries 100.0 & \bfseries 0.0 \\
        &  & Ours & 28.500001057982438 & 3.6634758128654634 & \bfseries 0.27077751850336074 & \bfseries 0.003795094831243368 & \bfseries 100.0 & \bfseries 0.0 & \bfseries 59.00000214576721 & \bfseries 3.0779357900923654 & \bfseries 0.27330570541418614 & \bfseries 0.004909022473389803 & \bfseries 100.0 & \bfseries 0.0 \\
        \cline{2-15}
        & \multirow[t]{7}{*}{DINOv2} & Random & 6.500000096857546 & 7.451598314743411 & 1.8354389482774558 & 0.22114214120552025 & 0.0 & 0.0 & 6.500000096857546 & 7.451598314743411 & 1.8139781839845945 & 0.20444725773313535 & 0.0 & 0.0 \\
        &  & LocalCKA~\cite{Maniparambil:2024:DVL} & 37.50000052154064 & 28.814287695238885 & 0.6419530018811201 & 0.1737451364368738 & 20.0 & 41.03913408340616 & 18.500000275671482 & 29.249381708103538 & 0.5300809075686436 & 0.1601697507371285 & 5.0 & 22.3606797749979 \\
        &  & OT~\cite{peyre2016gromov} & 30.000000558793538 & 13.764943952467299 & 1.3481522361196867 & 0.4570346187988009 & 0.0 & 0.0 & 33.50000027567148 & 19.808291269623112 & 1.3109703036940028 & 0.5472133884433561 & 0.0 & 0.0 \\
        &  & FAQ~\cite{vogelstein2015faq} & 37.50000111758709 & 21.244195063668812 & 0.6692816152685866 & 0.20670049092112167 & 0.0 & 0.0 & 38.00000071525574 & 29.841688109602934 & 0.5461875364304327 & 0.22656515756556306 & 0.0 & 0.0 \\
        &  & MPOpt~\cite{hutschenreiter2021fusion} & \bfseries 73.5000005364418 & \bfseries 17.851728883816058 & \bfseries 0.456536242266447 & \bfseries 0.012376669518300923 & 95.0 & 22.3606797749979 & 94.00000005960464 & 18.467610154073547 & 0.3250337641736231 & 0.023686823408301753 & 90.0 & 30.779350562554622 \\
        &  & Gurobi~\cite{gurobi} & 69.50000047683716 & 24.165003315452655 & \bfseries 0.45581642399271394 & \bfseries 0.011363459470699583 & \bfseries 100.0 & \bfseries 0.0 & \bfseries 100.0 & \bfseries 0.0 & \bfseries 0.31907762665707423 & \bfseries 0.01282123914296421 & \bfseries 100.0 & \bfseries 0.0 \\
        &  & Ours & 69.50000047683716 & 24.165003315452655 & \bfseries 0.45581642399271394 & \bfseries 0.011363459470699583 & \bfseries 100.0 & \bfseries 0.0 & \bfseries 100.0 & \bfseries 0.0 & \bfseries 0.31907762665707423 & \bfseries 0.01282123914296421 & \bfseries 100.0 & \bfseries 0.0 \\
        \cline{1-15} \cline{2-9}
        \multirow[c]{21}{*}{\rotatebox[origin=c]{90}{CINIC-10}} & \multirow[t]{7}{*}{CLIP} & Random & 6.500000096857546 & 7.451598314743411 & 1.8589628080639826 & 0.21898821718596317 & 0.0 & 0.0 & 6.500000096857546 & 7.451598314743411 & 1.834483391499101 & 0.2045981534105816 & 0.0 & 0.0 \\
        &  & LocalCKA~\cite{Maniparambil:2024:DVL} & 30.00000119209289 & 0.0 & 0.35553714229817385 & 0.11023704223306952 & 0.0 & 0.0 & 4.000000059604644 & 5.02624697439726 & 0.791845948777981 & 0.06085799485886409 & 0.0 & 0.0 \\
        &  & OT~\cite{peyre2016gromov} & 49.50000002980232 & 2.236067844219752 & 0.22895700986757705 & 0.008357773205438832 & 95.0 & 22.3606797749979 & 2.000000029802322 & 4.1039134694936905 & 0.9419153960537823 & 0.08525088268729847 & 0.0 & 0.0 \\
        &  & FAQ~\cite{vogelstein2015faq} & 30.500001162290566 & 2.236067844219754 & 0.26587616544615517 & 0.0083509538239828 & 0.0 & 0.0 & 0.0 & 0.0 & 0.6879560381016387 & 0.01881069375569753 & 0.0 & 0.0 \\
        &  & MPOpt~\cite{hutschenreiter2021fusion} & 0.0 & 0.0 & 0.70781341498288 & 0.004352622696188845 & 0.0 & 0.0 & 0.0 & 0.0 & 0.6491599203675993 & 0.010076465867738933 & 0.0 & 0.0 \\
        &  & Gurobi~\cite{gurobi} & \bfseries 50.0 & \bfseries 0.0 & \bfseries 0.22709527427121956 & \bfseries 0.001189002743407238 & \bfseries 100.0 & \bfseries 0.0 & \bfseries 80.0000011920929 & \bfseries 0.0 & \bfseries 0.38378149356117636 & \bfseries 0.002814817668735332 & \bfseries 100.0 & \bfseries 0.0 \\
        &  & Ours & \bfseries 50.0 & \bfseries 0.0 & \bfseries 0.22709527427121956 & \bfseries 0.001189002743407238 & \bfseries 100.0 & \bfseries 0.0 & \bfseries 80.0000011920929 & \bfseries 0.0 & \bfseries 0.38378149356117636 & \bfseries 0.002814817668735332 & \bfseries 100.0 & \bfseries 0.0 \\
        \cline{2-15}
        & \multirow[t]{7}{*}{DeiT} & Random & 6.500000096857546 & 7.451598314743411 & 1.8254873281204564 & 0.2282629191050487 & 0.0 & 0.0 & 6.500000096857546 & 7.451598314743411 & 1.8046952043632605 & 0.24404686061706965 & 0.0 & 0.0 \\
        &  & LocalCKA~\cite{Maniparambil:2024:DVL} & \bfseries 67.9999989271164 & \bfseries 8.94427137687901 & 0.3035261672702497 & 0.007541866184721837 & 0.0 & 0.0 & 0.0 & 0.0 & 0.6960680394488519 & 0.0035550467859165767 & 0.0 & 0.0 \\
        &  & OT~\cite{peyre2016gromov} & 20.00000029802322 & 0.0 & 0.4299090502840186 & 0.002273186599931137 & 0.0 & 0.0 & 0.0 & 0.0 & 0.9545607155754077 & 0.002981856328113909 & 0.0 & 0.0 \\
        &  & FAQ~\cite{vogelstein2015faq} & 55.500001311302185 & 5.1041790722712355 & 0.3484131785610666 & 0.023135816962943645 & 0.0 & 0.0 & 0.0 & 0.0 & 0.7053593343864628 & 0.0070831554554586255 & 0.0 & 0.0 \\
        &  & MPOpt~\cite{hutschenreiter2021fusion} & 0.0 & 0.0 & 0.7578251557686724 & 0.002646079226512082 & 0.0 & 0.0 & 0.0 & 0.0 & 0.695110200227142 & 0.004884770706127271 & 0.0 & 0.0 \\
        &  & Gurobi~\cite{gurobi} & 10.00000014901161 & 0.0 & \bfseries 0.26542529032294143 & \bfseries 0.0014365994080424097 & \bfseries 100.0 & \bfseries 0.0 & \bfseries 40.00000059604645 & \bfseries 0.0 & \bfseries 0.28098981330720696 & \bfseries 0.0012022378294992758 & \bfseries 100.0 & \bfseries 0.0 \\
        &  & Ours & 10.00000014901161 & 0.0 & \bfseries 0.26542529032294143 & \bfseries 0.0014365994080424097 & \bfseries 100.0 & \bfseries 0.0 & \bfseries 40.00000059604645 & \bfseries 0.0 & \bfseries 0.28098981330720696 & \bfseries 0.0012022378294992758 & \bfseries 100.0 & \bfseries 0.0 \\
        \cline{2-15}
        & \multirow[t]{7}{*}{DINOv2} & Random & 6.500000096857546 & 7.451598314743411 & 1.8374338713293032 & 0.24154880969944703 & 0.0 & 0.0 & 6.500000096857546 & 7.451598314743411 & 1.8087152539963731 & 0.24826427547927074 & 0.0 & 0.0 \\
        &  & LocalCKA~\cite{Maniparambil:2024:DVL} & 52.49999910593033 & 31.09831555275118 & 0.5130375311481482 & 0.16030700211198506 & 10.0 & 30.779350562554622 & 57.0000022649765 & 13.416408398118888 & 0.5796212058820734 & 0.10073552945758696 & 0.0 & 0.0 \\
        &  & OT~\cite{peyre2016gromov} & 77.5000011920929 & 6.386663540107756 & 0.4224781386137894 & 0.004525868906055283 & 90.0 & 30.779350562554622 & 15.500000342726704 & 7.591546882981429 & 1.4788303344270435 & 0.01511804161113689 & 0.0 & 0.0 \\
        &  & FAQ~\cite{vogelstein2015faq} & 31.000001132488244 & 4.472135688439506 & 0.5937087880708296 & 0.022838697187034386 & 0.0 & 0.0 & 29.500001147389405 & 2.2360681774198454 & 0.600222489267817 & 0.05457427937144293 & 0.0 & 0.0 \\
        &  & MPOpt~\cite{hutschenreiter2021fusion} & \bfseries 79.00000095367432 & \bfseries 3.0779357900923667 & \bfseries 0.4214813143614965 & \bfseries 0.002229764077478569 & \bfseries 100.0 & \bfseries 0.0 & 47.00000010430813 & 46.00915237041141 & 0.6133723304355289 & 0.15277834558820155 & 40.0 & 50.262468995003466 \\
        &  & Gurobi~\cite{gurobi} & \bfseries 79.00000095367432 & \bfseries 3.0779357900923667 & \bfseries 0.4214813143614965 & \bfseries 0.002229764077478569 & \bfseries 100.0 & \bfseries 0.0 & \bfseries 100.0 & \bfseries 0.0 & \bfseries 0.4586129193867702 & \bfseries 0.0039197635157094644 & \bfseries 100.0 & \bfseries 0.0 \\
        &  & Ours & \bfseries 79.00000095367432 & \bfseries 3.0779357900923667 & \bfseries 0.4214813143614965 & \bfseries 0.002229764077478569 & \bfseries 100.0 & \bfseries 0.0 & \bfseries 100.0 & \bfseries 0.0 & \bfseries 0.4586129193867702 & \bfseries 0.0039197635157094644 & \bfseries 100.0 & \bfseries 0.0 \\
        \bottomrule
    \end{tabularx}
    }
    \caption{\textbf{Vision-language alignment on CIFAR-10 and CINIC-10.} Our QAP solver achieves predominantly favourable matching accuracy, cost and optimality guarantees -- even in comparison to proprietary solvers (Gurobi). This holds across two datasets and three pre-trained models: CLIP, DeiT and DINOv2.}
    \label{tab:small_benchmark_solvers_app}
\end{table*}

\section{Optimal transport as QAP solver}
\label{appendix:ot_doubly_stochastic}
We show that optimal transport with uniform source and target probability distributions is equivalent to relaxing the QAP with doubly stochastic matrices. In our experiments, we use this equivalence to compare to the solutions from optimal transport solvers.

For uniform probability distributions $\mathbf{p} = \mathbf{q} = \frac{1}{N} \mathbbm{1}$, let
\begin{align}
    \mathbf{T^*} \in \argmin_{\substack{\mathbf{T} \in [0, 1]^{N \times N}\\\mathbf{T} \mathbbm{1} = \mathbf{p}\\\mathbf{T}^T \mathbbm{1} = \mathbf{q}}} \mathbf{L}_{i j k l} \mathbf{T}_{i j} \mathbf{T}_{k l}
\end{align}
be an optimal transport matrix for the four axes tensor $\mathbf{L} \in \R^{N \times N \times N \times N}$. Then, for any other transport matrix $\mathbf{T} \in [0, 1]^{N \times N}$ with $\mathbf{T} \mathbbm{1} = \mathbf{p}$ and $\mathbf{T}^T\mathbbm{1} = \mathbf{q}$, it holds that
\begin{align}
    \mathbf{L}_{i j k l} \mathbf{T}_{i j} \mathbf{T}_{k l} \geq \mathbf{L}_{i j k l} \mathbf{T^*}_{i j} \mathbf{T^*}_{k l}.
\end{align}
Now, we can define our optimal stochastic matrix $\mathbf{S^*} \in [0, 1]^{N \times N}$ by $\mathbf{S^*} = N \mathbf{T^*}$. This is indeed a stochastic matrix as $N \mathbf{T^*} \geq 0$, $\mathbf{S^*} \mathbbm{1} = N \mathbf{T^*} \mathbbm{1} = N \mathbf{p} = \mathbbm{1}$ and $(\mathbf{S^*})^T \mathbbm{1} = N \mathbf{T^*} \mathbbm{1} = N \mathbf{q} = \mathbbm{1}$. Moreover, this stochastic matrix is optimal, because for every other stochastic matrix $\mathbf{S} \in [0, 1]^{N \times N}$,
\begin{align}
    \mathbf{L}_{i j k l} \mathbf{S}_{i j} \mathbf{S}_{k l} = &N ^ 2 \mathbf{L}_{i j k l} \frac{\mathbf{S}_{i j}}{N} \frac{\mathbf{S}_{k l}}{N} \geq \\
    &\geq N ^ 2 \mathbf{L}_{i j k l} \mathbf{T^*}_{i j} \mathbf{T^*}_{k l} = \mathbf{L}_{i j k l} \mathbf{S^*}_{i j} \mathbf{S^*}_{k l}.\nonumber
\end{align}
Here, we use the fact that $\frac{\mathbf{S}}{N}$ is a valid transport matrix as $\frac{\mathbf{S}}{N} \mathbbm{1} = \frac{1}{N} \mathbbm{1} = \mathbf{p}$ and $\frac{\mathbf{S}^T}{N} \mathbbm{1} = \frac{1}{N} \mathbbm{1} = \mathbf{q}$. The other direction can be derived in a similar way by dividing the doubly stochastic matrix by $N$.

\vfill\null

\end{document}